\documentclass[letterpaper]{article} 
\usepackage{aaai2026}  
\usepackage{times}  
\usepackage{helvet}  
\usepackage{courier}  
\usepackage[hyphens]{url}  
\usepackage{graphicx} 
\usepackage{natbib}  
\usepackage{caption} 
\usepackage{algorithm}

\usepackage{amsmath}
\usepackage{xcolor}
\usepackage{amssymb} 

\usepackage{amsmath}
\usepackage[noend]{algpseudocode} 

\usepackage{booktabs}  
\usepackage{siunitx}   
\usepackage{multirow}  
\usepackage{makecell} 
\usepackage{color} 
\usepackage{colortbl} 
\usepackage{pifont} 
\usepackage{calc} 

\usepackage{newfloat}
\usepackage{listings}
\DeclareCaptionStyle{ruled}{labelfont=normalfont,labelsep=colon,strut=off} 
\floatstyle{ruled}
\newfloat{listing}{tb}{lst}{}
\floatname{listing}{Listing}
\title{Unsupervised Ultra-High-Resolution UAV Low-Light Image Enhancement: \\ A Benchmark, Metric and Framework}
\author{
    Wei Lu\textsuperscript{\rm 1}, Lingyu Zhu\textsuperscript{\rm 2}, Si-Bao Chen\textsuperscript{\rm 1}\thanks{Si-Bao Chen is corresponding author.}
}
\affiliations{
    \textsuperscript{\rm 1}School of Computer Science and Technology, Anhui University\\
    \textsuperscript{\rm 2}Computer Science, City University of Hong Kong \\
    luwei\_ahu@qq.com, lingyzhu-c@my.cityu.edu.hk, sbchen@ahu.edu.cn
}

\usepackage{bibentry}

\begin{document}

\maketitle

\begin{abstract}
Low-light conditions significantly degrade Unmanned Aerial Vehicles (UAVs) performance in critical applications. Existing Low-light Image Enhancement (LIE) methods struggle with the unique challenges of aerial imagery, including Ultra-High Resolution (UHR), lack of paired data, severe non-uniform illumination, and deployment constraints. To address these issues, we propose three key contributions. First, we present U3D, the first unsupervised UHR UAV dataset for LIE, with a unified evaluation toolkit. Second, we introduce the Edge Efficiency Index (EEI), a novel metric balancing perceptual quality with key deployment factors: speed, resolution, model complexity, and memory footprint. Third, we develop U3LIE, an efficient framework with two training-only designs—Adaptive Pre-enhancement Augmentation (APA) for input normalization and a Luminance Interval Loss ($\mathcal{L}_{int}$) for exposure control. U3LIE achieves SOTA results, processing 4K images at 23.8 FPS on a single GPU, making it ideal for real-time on-board deployment. In summary, these contributions provide a holistic solution (dataset, metric, and method) for advancing robust 24/7 UAV vision.  The code and datasets are available at https://github.com/lwCVer/U3D$\_$Toolkit.
\end{abstract}


\section{Introduction}	\label{sec_Introduction}
\begin{figure}[t]
	\centering
	\includegraphics[width=\linewidth]{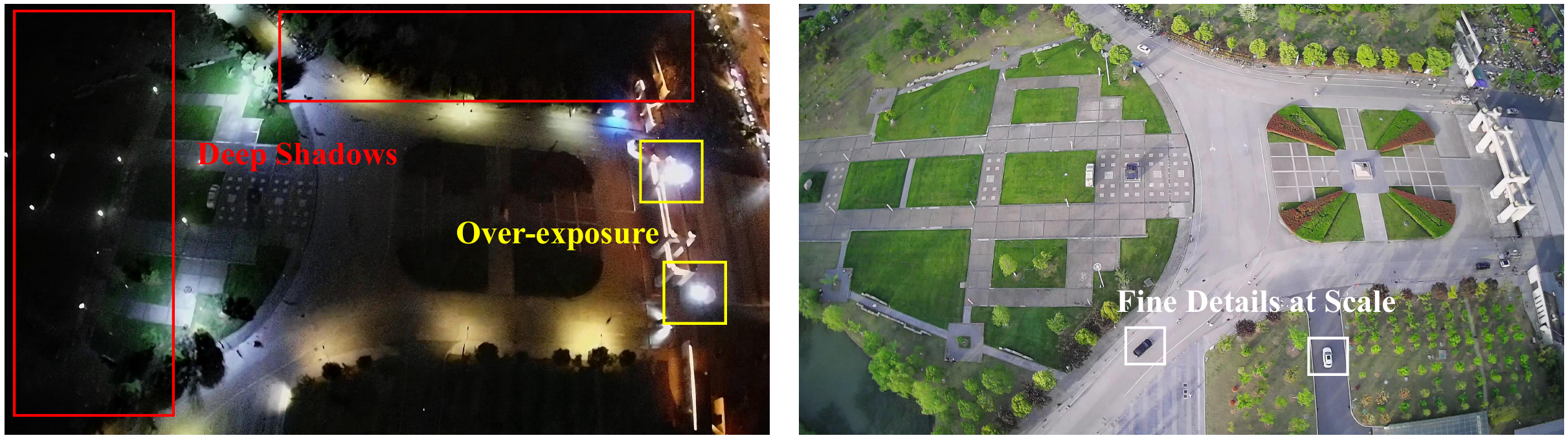} 
	\caption{Examples from our U3D dataset. Left: low-light night image. Right: non-paired daytime image. The UAV low-light scenes feature: critical small-scale details, unpaired data, deep shadows, and over-exposed lights.
	}
	\label{fig:u3d_examples}
\end{figure}

Unmanned Aerial Vehicles (UAVs) are increasingly deployed in mission-critical tasks such as disaster response and urban surveillance. However, their effectiveness is hindered in low-light environments, where captured imagery often suffers from noise, low contrast, and loss of detail. These degradations impair downstream vision tasks and threaten system reliability, making Low-light Image Enhancement (LIE) a critical technology for ensuring robust UAV operations 24/7.
Unfortunately, existing LIE methods, mostly developed for ground-level imagery, generalize poorly to aerial views due to four fundamental challenges (Fig.~\ref{fig:u3d_examples}):
(1) UAVs typically operate at high altitudes, requiring \textbf{Ultra-High Resolution (UHR)} to preserve fine-scale details of small ground objects;
(2) The dynamic nature of UAV flight makes it infeasible to collect aligned low/normal-light image pairs, necessitating \textbf{unsupervised learning} approaches;
(3) Aerial night scenes exhibit \textbf{extreme non-uniform illumination}, with over-exposed city lights and under-exposed dark zones coexisting in a single frame;
(4) \textbf{On-board deployment} demands models with low latency, memory footprint, and computational complexity for real-time inference on resource-limited platforms.

These combined challenges underscore the inadequacy of simply adapting ground-level LIE methods for UAV applications. This necessitates a specialized solution that jointly addresses data availability, aerial-specific visual characteristics, and deployment constraints.

To this end, we present a three-part solution designed to systematically address these challenges.
\textbf{First, we address the foundational problem of data and evaluation platforms.} We construct and release the \textbf{U3D (Unsupervised UHR UAV Dataset)}, the first dataset specifically designed for unsupervised LIE on UAVs. U3D contains 1,000 real-world nocturnal aerial images at 3840$\times$2160 resolution and 1,000 non-paired daytime images of the same size, providing a solid data foundation for the development and validation of unsupervised models. Furthermore, we develop a unified \textbf{U3D toolkit} that integrates 12 representative methods to provide a fair and reproducible benchmark for future research.
\textbf{Second, to more accurately assess the overall performance of algorithms in real-world UAV applications, we propose a new evaluation metric: the Edge Efficiency Index (EEI).} Unlike traditional metrics that focus solely on visual quality, EEI combines enhancement performance with efficiency-related factors, including inference speed, image resolution, parameter count, and computational complexity. The index is designed to provide a more comprehensive measure that is closely aligned with practical deployment needs, thereby guiding algorithm design toward a better balance between quality and efficiency.
\textbf{Finally, to tackle the dual challenges of non-uniform illumination and on-board deployment, we propose the U3LIE framework.} U3LIE (Unsupervised UHR UAV LIE) incorporates two main innovations:
\textbf{1) Adaptive Pre-enhancement Augmentation (APA):} This is a data augmentation technique used during the training phase. It preprocesses extremely dark input images through a pipeline that includes adaptive gamma correction and local contrast enhancement. The goal is to normalize the luminance distribution of the input, simplifying the learning manifold for the network, thereby mitigating over-exposure in bright areas and noise amplification in dark areas during the final enhancement.
\textbf{2) Luminance Interval Loss ($\mathcal{L}_{int}$):} It is designed to address the critical trade-off between enhancing dark regions and suppressing bright regions. Unlike conventional losses that aim for a single target brightness, our $\mathcal{L}_{int}$ enforces a desirable luminance interval. It simultaneously penalizes local regions that are too dark (falling below a lower threshold) and those that are too bright (exceeding an upper threshold). This mechanism guides the network to restore details in shadows without causing over-exposure in highlights, which is suitable for high dynamic range UAV scenes.

Experimental results on the U3D dataset show that U3LIE achieves an excellent balance between enhancement performance and operational efficiency. Notably, when processing 4K (3840$\times$2160) resolution images, U3LIE runs at 23.8 Frames Per Second (FPS), 3$\times$ faster than the competing unsupervised methods such as RUAS (7.6 FPS), while also demanding fewer parameters and FLOPs. This demonstrates its strong potential for practical on-board deployment.

Our main contributions are summarized as follows:
\begin{itemize}
	\item We construct and release the first unsupervised UHR UAV dataset (U3D) for LIE, along with its accompanying unified evaluation U3D toolkit.
	\item We propose a novel Edge Efficiency Index (EEI) for the comprehensive evaluation of algorithm quality and deployment efficiency in UAV contexts.
	\item We design the U3LIE framework, which effectively improves the performance-efficiency balance for UAVs through the APA training strategy and a novel $\mathcal{L}_{int}$.
\end{itemize}

\section{Related Work}		\label{Related_Work}
Deep learning-based LIE~\cite{wei2018deep} has evolved through several paradigms. Supervised methods, which rely on paired low/normal-light datasets, are often inspired by Retinex theory~\cite{zhang2019kindling, wu2022uretinexnet} or employ end-to-end architectures that learn direct mappings~\cite{lv2018mbllen, chen2018learning, wang2018gladnet, wang2022lowlight}. However, acquiring aligned data for dynamic UAVs is challenging. To overcome this, unsupervised approaches leverage unpaired data, often using GANs~\cite{zhu2017unpaired, jiang2021enlightengan, park2020contrastive, yang2020from, zhu2024unrolled, zhu2024temporally}. A particularly flexible and lightweight category is zero-reference learning, pioneered by Zero-DCE~\cite{guo2020zeroreference}. This approach trains models guided only by non-reference losses, yielding a family of efficient methods well-suited for on-board deployment~\cite{liu2021retinexinspired, ma2022toward, li2022learning, liang2023iterative, shi2024zeroig} and forming the basis of our work.

However, progress in the aerial domain is constrained by inadequate benchmarks. Existing datasets are either low-resolution~\cite{wei2018deep, chen2018learning}, designed for supervised ground-level scenes~\cite{wang2023ultrahighdefinition}, or lack the top-down perspective and UHR required for UAVs~\cite{yu2020bdd100k, lohia2019exdark}. No benchmark is tailored for unsupervised UHR UAV enhancement. Crucially, standard evaluation metrics overlook the practical requirements of UAV applications. Metrics focused on fidelity (e.g., PSNR/SSIM~\cite{wang2004image}) or perceptual quality (e.g., NIQE~\cite{mittal2012making}, BRISQUE~\cite{mittal2012noreference}, and PI~\cite{blau2018perceptiondistortion}) focus on visual results and disregard edge efficiency—a critical factor for on-board systems.

\section{U3D Dataset and Toolkit} \label{sec:dataset_toolkit}
To address data scarcity and the lack of standardized evaluation in UAV LIE, we introduce the \textbf{U}nsupervised \textbf{U}AV \textbf{U}HR \textbf{D}ataset (\textbf{U3D}) and its evaluation \textbf{U3D toolkit}.

\begin{figure}[t] 
	\centering
	\includegraphics[width=\linewidth]{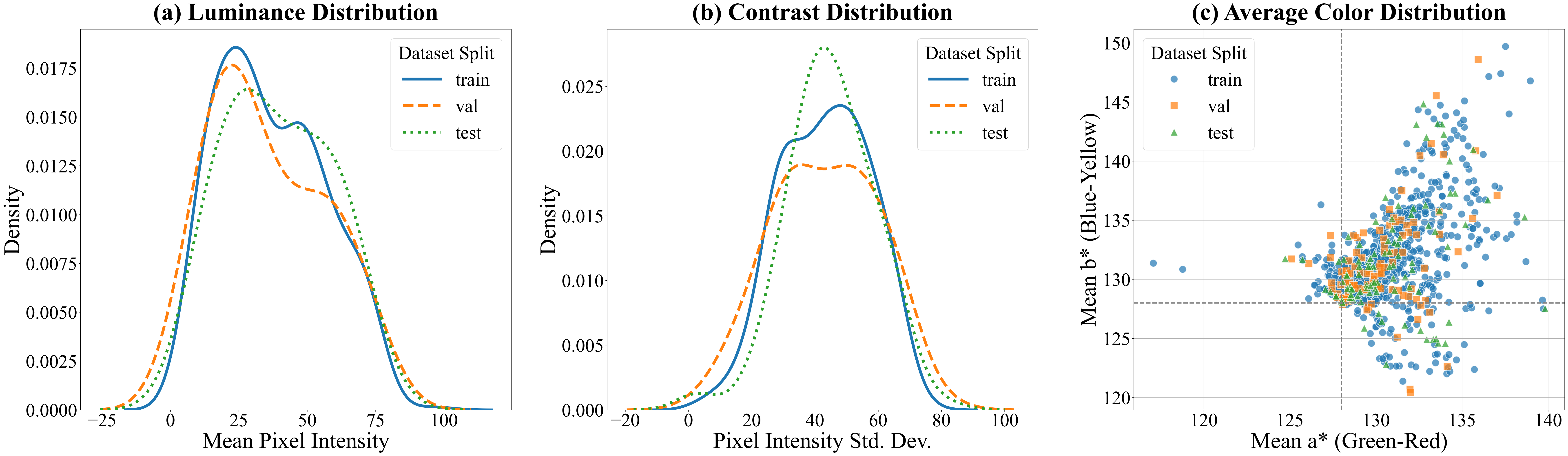} 
	\caption{Statistical analysis of the U3D dataset splits.
		(a) Luminance distribution, (b) Contrast distribution, and (c) Average color distribution in CIELAB space. The high degree of overlap between the train, validation, and test sets demonstrates a consistent and unbiased data partitioning.
	}
	\label{fig:dataset_stats}
\end{figure}

\subsection{The U3D Dataset}
U3D is an unpaired dataset designed for unsupervised learning, given the inherent difficulty in capturing perfectly aligned image pairs from dynamic UAV flights. The images were captured in real-world scenarios across diverse locations. It contains 1,000 UHR (3840$\times$2160) low-light images and 1,000 non-paired daytime images, covering diverse urban, residential, and natural scenes. Captured with a professional gimbal-stabilized UAV at altitudes from 30-120m, the images are direct camera outputs, preserving authentic sensor noise and artifacts.
To ensure fair and reproducible comparisons, U3D is partitioned into training, validation, and test sets (8:1:1). As shown in Fig.~\ref{fig:dataset_stats}, the statistical properties across these splits are consistent, confirming the absence of partitioning bias and supporting fair model evaluation.

\subsection{The U3D Toolkit}
To facilitate standardized and reproducible research, we develop U3D toolkit, a comprehensive and modular evaluation platform. It decouples the research pipeline---including data loading, model architectures, loss functions, and training strategies---into interchangeable modules. This allows researchers to easily integrate and test new components. To provide a robust baseline for future work, the toolkit already integrates 12 representative models, which are used for comparison in experiments. By providing this unified platform, the U3D toolkit significantly facilitates broader participation for conducting rigorous experiments in this domain.

\section{Edge Efficiency Index (EEI)}
\label{sec:eei}

Traditional no-reference image quality assessment (NR-IQA) metrics like NIQE \cite{mittal2012making} evaluate perceptual quality but ignore computational efficiency---a critical factor for on-board UAV deployment. 
We introduce EEI to evaluate LIE models for resource-constrained scenarios by balancing perceptual quality with computational cost. EEI systematically balances perceptual quality with a multi-dimensional measure of computational cost. A lower score indicates a more favorable quality-efficiency trade-off. The EEI is formulated as:
\begin{equation}
	\label{eq:eei_main}
	\text{EEI} = \mathcal{Q}_{\text{PI}} \cdot \mathcal{E}_{\text{norm}},
\end{equation}
where $\mathcal{Q}_{\text{PI}}$ is a perceptual quality score and $\mathcal{E}_{\text{norm}}$ is a normalized multi-faceted efficiency factor.

\paragraph{Perceptual Quality ($\mathcal{Q}_{\text{PI}}$).}
For the quality component, we use the PI, defined as the average of NIQE and BRISQUE scores, to measure image naturalness and artifacts.

\paragraph{Normalized Efficiency ($\mathcal{E}_{\text{norm}}$).}
The efficiency factor $\mathcal{E}_{\text{norm}}$ aggregates three cost components: runtime, complexity, and memory. To ensure fair comparisons across different hardware, all components are normalized against a standard baseline established by running MobileNetV2 \cite{sandler2018mobilenetv} on the target device. This one-time calibration is performed at a fixed base resolution $R_{base}$ of 3840$\times$2160 and yields reference values for inference time ($T_{ref}$), FLOPs ($F_{base}$), parameters ($P_{base}$), and memory usage ($M_{base}$).

For a model under evaluation with performance metrics $T_{model}, F_{model}, P_{model}, M_{model}$ at resolution $R_{model}$, we define three normalized cost factors: \\
\textbf{Time Factor ($\mathcal{E}_{\text{time}}$):} $\left( T_{model} / T_{ref} \right) \cdot \left( R_{base} / R_{model} \right)$, which accounts for both speed and processing resolution. \\
\textbf{Complexity Factor ($\mathcal{E}_{\text{comp}}$):} An equal-weighted average of normalized FLOPs and parameters: $0.5 \cdot (F_{model}/F_{base} + P_{model}/P_{base})$. \\
\textbf{Resource Factor ($\mathcal{E}_{\text{rsrc}}$):} Normalized memory usage: $M_{model} / M_{base}$.

The final efficiency factor, $\mathcal{E}_{\text{norm}}$, is a weighted sum of these factors. To robustly handle cases where profiling FLOPs/parameters fails due to out-of-memory (OOM) errors (common for large models on UHR inputs), we employ a pragmatic adaptive weighting scheme:
\begin{equation}
	\mathcal{E}_{\text{norm}} =
	\begin{cases} 
		w'_t \mathcal{E}_{\text{time}} + w'_r \mathcal{E}_{\text{rsrc}}, & \hspace{-6mm} \text{if } \mathcal{E}_{\text{comp}} \hspace{2mm} \text{OOM}, \\
		w_t \mathcal{E}_{\text{time}} + w_c \mathcal{E}_{\text{comp}} + w_r \mathcal{E}_{\text{rsrc}}, & \text{otherwise}.
	\end{cases}
	\label{eq:eei_final}
\end{equation}
We set the default weights to $w_t=0.8, w_c=0.1, w_r=0.1$ to prioritize inference speed (runtime), which is often the most critical bottleneck for real-time UAV applications such as obstacle avoidance or target tracking. 
The EEI is a flexible framework; weights can be adapted to specific constraints. For instance, on devices with limited GPU memory, one might increase $w_r$ to penalize high memory usage. This adaptability allows the EEI to serve as a versatile tool for evaluating algorithm suitability across a spectrum of real-world, resource-constrained use cases.

\begin{figure*}[t]
	\centering
	\includegraphics[width=0.92\linewidth]{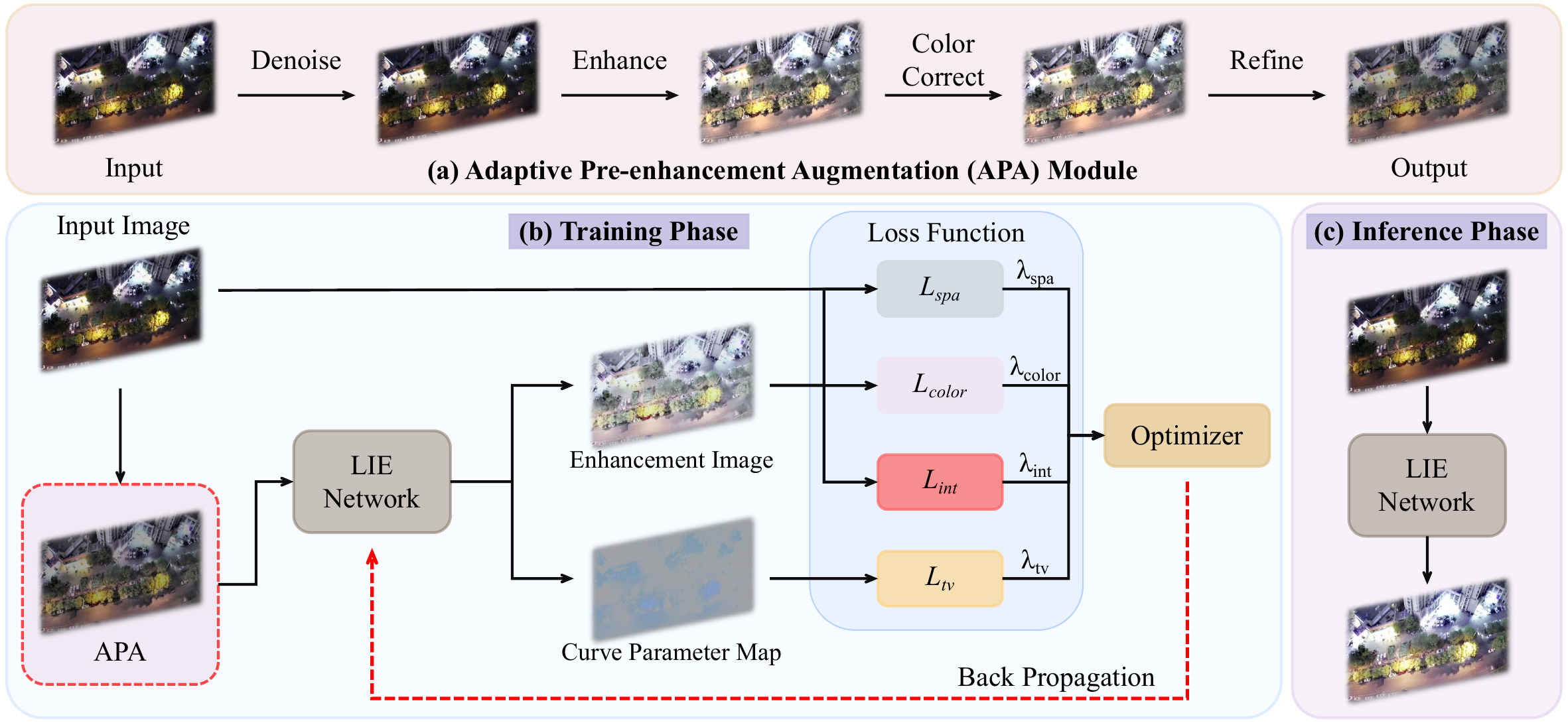}
	\caption{
		The U3LIE framework. During training (b), a training-only APA module (a) preprocesses input for the network, which is optimized with a loss suite including our $\mathcal{L}_{int}$. During inference (c), the input is processed directly by the trained network.} \label{fig:framework}
\end{figure*}

\section{The Proposed U3LIE Framework} \label{sec:u3lie_framework}
This section presents our proposed unsupervised UHR UAV LIE framework, \textbf{U3LIE}. The pipeline includes three components: (1) Adaptive Pre-enhancement Augmentation (APA), applied only during training; (2) the enhancement network; and (3) the luminance interval loss $\mathcal{L}_{int}$, which acts as a key regularization constraint. U3LIE operates in an end-to-end manner, enabling efficient enhancement without requiring supervision or priors during inference.

\subsection{Lightweight Enhancement Network} \label{ssec:network}
To ensure on-board efficiency, our design philosophy prioritizes computational performance without compromising visual quality. This principle guides our choice of a highly lightweight enhancement network based on the curve-based methodology of DCE++ \cite{li2022learning}. This approach contrasts with the prevailing trend of developing larger models, which, despite their power, often remain impractical for resource-constrained UAVs.

The network learns a pixel-wise enhancement map $A$ to parameterize an iterative, higher-order curve that adjusts the input image's illumination. The final enhanced image is generated by applying this curve. To further optimize for our specific use case, we modify the original DCE++ architecture by: \textbf{(1) removing its downsampling/upsampling stages} to process inputs at their native resolution, and \textbf{(2) reducing the network's channel depth} from 32 to 8. 

\begin{table*}[ht]\centering \small
	\sisetup{table-format=2.2, round-mode=places, round-precision=2, table-space-text-post={***} }
	\begin{tabular}{
			lccrr
			S[table-format=2.2] 
			S[table-format=2.2] 
			S[table-format=1.2]
			S[table-format=2.2] 
			S[table-format=2.2] 
			S[table-format=4.2] } \specialrule{0.8pt}{0pt}{0pt}
		\rowcolor[rgb]{0.9,0.9,0.9} \textbf{Method} 
		& \textbf{Type} & \textbf{Venue}         
		& \textbf{{\#P  $\downarrow$}}
		& \textbf{{FLOPs  $\downarrow$}}
		& \textbf{{Memory $\downarrow$}}
		& \textbf{{FPS $\uparrow$}}   & \textbf{{NIQE $\downarrow$}}   
		& \textbf{{BRISQUE $\downarrow$}}   & \textbf{{PI $\downarrow$}}                
		& \textbf{{EEI $\downarrow$}} \\ \hline \hline
		LLFormer     & Sup & AAAI'23   	&OOM	&OOM 	&5.06{*} 	& 0.06  & 4.68	& 32.73	& 18.70 & 6612.54 \\
		RetinexFormer& Sup & ICCV'23   	&OOM	&OOM 	&2.066{*}	& 0.33  & 4.44  & 31.63 & 18.04 & 1208.42 \\
		HVI          & Sup & CVPR'25   	&1.97M	&1029.00& 15.11 	& 0.56  & 4.94  & 35.46 & 20.20 & 742.42 \\ 
		MobileIE	 & Sup & ICCV'25   	&4.05K	&31.95	& 3.11 		& 26.95 & 4.65 	& 41.34	& 22.99	& 20.50 \\ \midrule
		CycleGAN     & Unp & ICCV'17   	&11.38M	&7197.0	& 14.03 	& 1.38  & 2.97  & 8.55  & 5.76  & 120.74 \\
		NeRCo        & Unp & ICCV'23   	&OOM	&OOM 	& 3.28{*} 	& 0.10  & 4.29  & 16.21 & 10.25 & 2180.89 \\ \midrule
		ZERO-DCE     & Uns & CVPR'20   	&79.42K &656.92 & 12.89 	& 7.76  & 3.89  & 39.43 & 21.66 & 69.29 \\
		RUAS         & Uns & CVPR'21   	&3.44K	&27.10  & 7.75  	& 7.64  & 4.83  & 38.62 & 21.73 & 57.91 \\
		DCE++        & Uns & PAMI'22   	&10.56K &83.57  & 17.06 	& 7.68  & 4.19  & 37.46 & 20.82 & 56.39 \\
		SCI          & Uns & CVPR'22   	&10.67K &1451.00& 6.92  	& 2.98  & 4.54  & 37.38 & 20.96 & 167.07 \\
		ZERO-IG      & Uns & CVPR'24   	&123.63K&4293.00& 15.78 	& 3.09  & 4.78  & 37.50 & 21.14 & 218.31 \\
		Iformer		 & Uns & ICLR'25 	&OOM	&OOM 	& 2.09{*} 	& 5.07  & 5.23  & 36.72 & 20.98 & 2620.69 \\ \hline
		\rowcolor[rgb]{0.9,0.9,0.9}
		\textbf{U3LIE (Ours)}&Uns &-	&1.32K	& 9.91 	& 3.23 		& 23.80 & 4.30	& 29.92	& 17.11	& 16.03
		\\ \specialrule{0.8pt}{0pt}{0pt}
	\end{tabular}
	\caption{
		Quantitative comparison with SOTA methods on the U3D dataset.
		All models are evaluated on 4K resolution images (3840$\times$2160). 
		``Sup'', ``Unp'', and ``Uns'' denote Supervised, Unpaired, and Unsupervised methods, respectively. For supervised (Sup) methods, we directly test using the officially provided pre-trained weights without any fine-tuning.
		\#P: Number of parameters (K: $10^3$, M: $10^6$). 
		FLOPs: Giga Floating-point operations per second (G: $10^9$).
		Memory: GPU memory usage in Gigabytes (GB).
		An asterisk (*) indicates that the model was evaluated using a patch-based approach due to memory constraints.
		OOM denotes an Out-of-Memory error on a single NVIDIA RTX 3090 GPU.
		$\downarrow$: Lower is better. $\uparrow$: Higher is better.
	} \label{table:sota_comparison}
\end{table*}

\subsection{Adaptive Pre-enhancement Augmentation (APA)} \label{ssec:apa}
Directly training on UAV images with extreme dynamic range is challenging. We introduce \textbf{APA}, a \textbf{training-only, one-time pre-processing} transform, $\mathcal{T}_{APA}$, that generates a moderately-enhanced augmented training set. The pipeline is designed as a two-stage process: first, an adaptive luminance and contrast boost to improve visibility in dark regions, followed by a perceptual correction stage to address color casts and suppress highlights, common artifacts in nocturnal aerial scenes. It simplifies the learning manifold by reducing severe luminance variations and color casts, thus preventing the network from producing over-exposed results or amplifying noise. The APA pipeline is a two-stage process designed to first enhance visibility and then refine perceptual quality, as defined mathematically below.

\paragraph{1. Adaptive Luminance and Contrast Boosting ($\mathcal{T}_{Lum}$).}
After an initial denoising step $\mathcal{F}_{Denoise}$ using a bilateral filter, the core luminance enhancement is performed in the YCrCb color space for independent control. We first compute the normalized mean luminance, denoted as $\hat{\mu}_Y$, by averaging the pixel values of the input's luminance channel $Y$ and normalizing by the maximum level $L_{max}$ (e.g., 255). This mean value is used to derive an adaptive gamma, $\gamma_{adapt}$, which adjusts the enhancement strength based on the image's overall darkness:
\begin{equation}
	\gamma_{adapt} = \text{clip}\left(\gamma_{base} - \kappa \cdot \ln(\hat{\mu}_Y + \epsilon), [\gamma_{min}, \gamma_{max}]\right).
\end{equation}
Here, $\gamma_{base}$ is a base gamma value, $\kappa$ is a sensitivity hyperparameter, and $\epsilon$ is a small constant for numerical stability. The luminance channel is then enhanced via this adaptive gamma correction, followed by Contrast Limited Adaptive Histogram Equalization (CLAHE)~\cite{pizer1987adaptive} to boost local details. The entire transformation on the luminance channel is:
\begin{equation}
	Y'(p) = \mathcal{F}_{CLAHE}\left(L_{max} \cdot \left(Y(p) / L_{max}\right)^{1/\gamma_{adapt}}\right).
\end{equation}
The final image from this stage, $I''$, is obtained by merging the enhanced $Y'$ with the original Cr and Cb channels and converting back to the RGB space.

\paragraph{2. Perceptual Color and Highlight Correction} ($\mathcal{T}_{Color}$ and $\mathcal{T}_{Final}$).
To address common color casts and preserve highlight details, we apply two subsequent correction steps. First, to correct reddish or greenish tints, the image $I''$ is converted to the CIEL*a*b* color space. The a* channel, which represents the green-red axis, is adjusted using a boost factor $\beta_{red}$:
\begin{equation}
	a'(p) = (a(p) - c_{mid}) \cdot \beta_{red} + c_{mid},
\end{equation}
where $c_{mid} = L_{max}/2$ is the neutral midpoint of the channel. This step yields the image $I'''$.

Finally, to restore color vibrancy and prevent over-exposure, $I'''$ is processed in the HSV space. The saturation (S) and value (V) channels are independently scaled by factors $\beta_{sat}$ and $\eta_{supp}$ respectively:
\begin{align}
	S'(p) &= \text{clip}(S(p) \cdot \beta_{sat}, [0, L_{max}]), \\
	V'(p) &= \text{clip}(V(p) \cdot \eta_{supp}, [0, L_{max}]). \label{eq:v_suppress}
\end{align}
The value suppression in Eq.(\ref{eq:v_suppress}) is crucial for suppressing over-exposed regions, a common artifact in nighttime aerial scenes with artificial lighting. The resulting image $I_{aug}$ serves as the normalized input for network training.

\begin{figure}[t]
	\centering
	\includegraphics[width=0.96\linewidth]{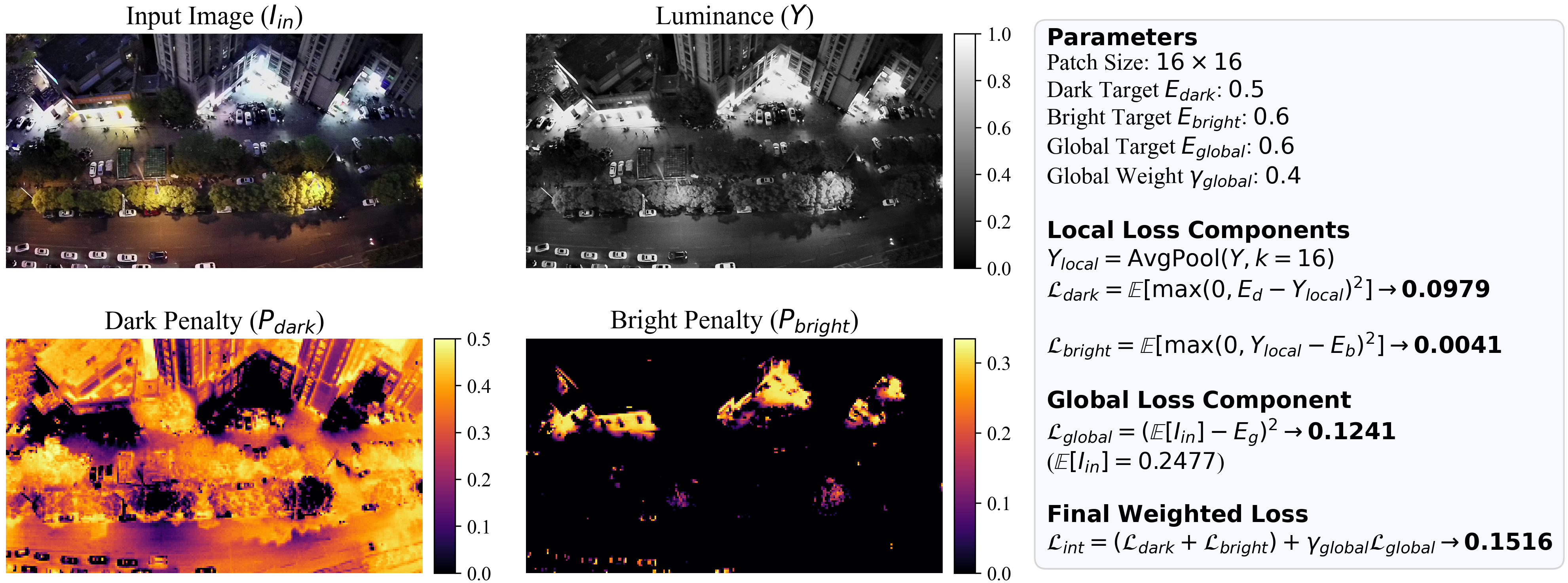} 
	\caption{Analysis of $\mathcal{L}_{int}$. Left: Visualization of penalty maps for dark/bright regions. Right: Quantitative summary.}
	\label{fig:lint_analysis}
\end{figure}

\subsection{Luminance Interval Loss ($\mathcal{L}_{int}$)}
\label{ssec:lint}
To overcome the limitations of single-target exposure losses, we introduce the Luminance Interval Loss ($\mathcal{L}_{int}$), which enables fine-grained control over luminance. Let $I_{enh}$ be partitioned into a set of non-overlapping patches $\{\mathcal{P}_i\}$. For each patch, we compute its mean luminance $Y_{\mathcal{P}_i} = \mathbb{E}_{p \in \mathcal{P}_i}[Y_{enh}(p)]$. The loss consists of three components:

\paragraph{1. Local Luminance Penalties.}
To enforce that patch luminances lie within a desired interval $[E_{dark}, E_{bright}]$, we define under- and over-exposure penalties:
\begin{align}
	\mathcal{L}_{local\_dark}(I_{enh}) &= \mathbb{E}_{i} \left[ \max(0, E_{dark} - Y_{\mathcal{P}_i})^2 \right], \\
	\mathcal{L}_{local\_bright}(I_{enh}) &= \mathbb{E}_{i} \left[ \max(0, Y_{\mathcal{P}_i} - E_{bright})^2 \right].
\end{align}

\paragraph{2. Global Brightness Constraint.}
To regulate the global image brightness, we regularize the global mean luminance $\bar{Y}_{global} = \mathbb{E}_{p \in \Omega}[Y_{enh}(p)]$ towards a target $E_{global}$:
\begin{equation}
	\mathcal{L}_{global}(I_{enh}) = (\bar{Y}_{global} - E_{global})^2.
\end{equation}
The complete $\mathcal{L}_{int}$ is a weighted sum of these components:
\begin{equation}
	\mathcal{L}_{int}(I_{enh}) = (\mathcal{L}_{local\_dark} + \mathcal{L}_{local\_bright}) + \gamma_{\text{global}} \mathcal{L}_{global}.
\end{equation}

Fig.~\ref{fig:lint_analysis} provides a visual breakdown of how these components operate on a sample image, demonstrating the loss's ability to identify and penalize both under- and over-exposed regions simultaneously.

\subsection{Total Loss Function}
The framework is trained end-to-end by minimizing a total loss function that combines our proposed $\mathcal{L}_{int}$ with a set of established non-reference losses for smoothness of the enhancement map $A$ ($\mathcal{L}_{tv}$), spatial coherence ($\mathcal{L}_{spa}$), and color constancy ($\mathcal{L}_{col}$). The total loss is formulated as:
\begin{equation}
	\begin{split}
		\mathcal{L}_{total} = & \lambda_{int}\mathcal{L}_{int}(I_{enh})  +  \lambda_{spa}\mathcal{L}_{spa}(I_{enh}, I_{in}) \\
		& + \lambda_{tv}\mathcal{L}_{tv}(A) + \lambda_{col}\mathcal{L}_{col}(I_{enh}),
	\end{split}
	\label{eq:total_loss}
\end{equation}
where $I_{enh} = \mathcal{C}^{(N)}(I_{in}, A)$ and $A = \mathcal{F}_{net}(\mathcal{T}_{APA}(I_{in}))$. The $\lambda$ terms are the corresponding weights.

\section{Experiments} \label{sec:experiments}

\begin{figure}[t]
	\centering
	\includegraphics[width=\linewidth]{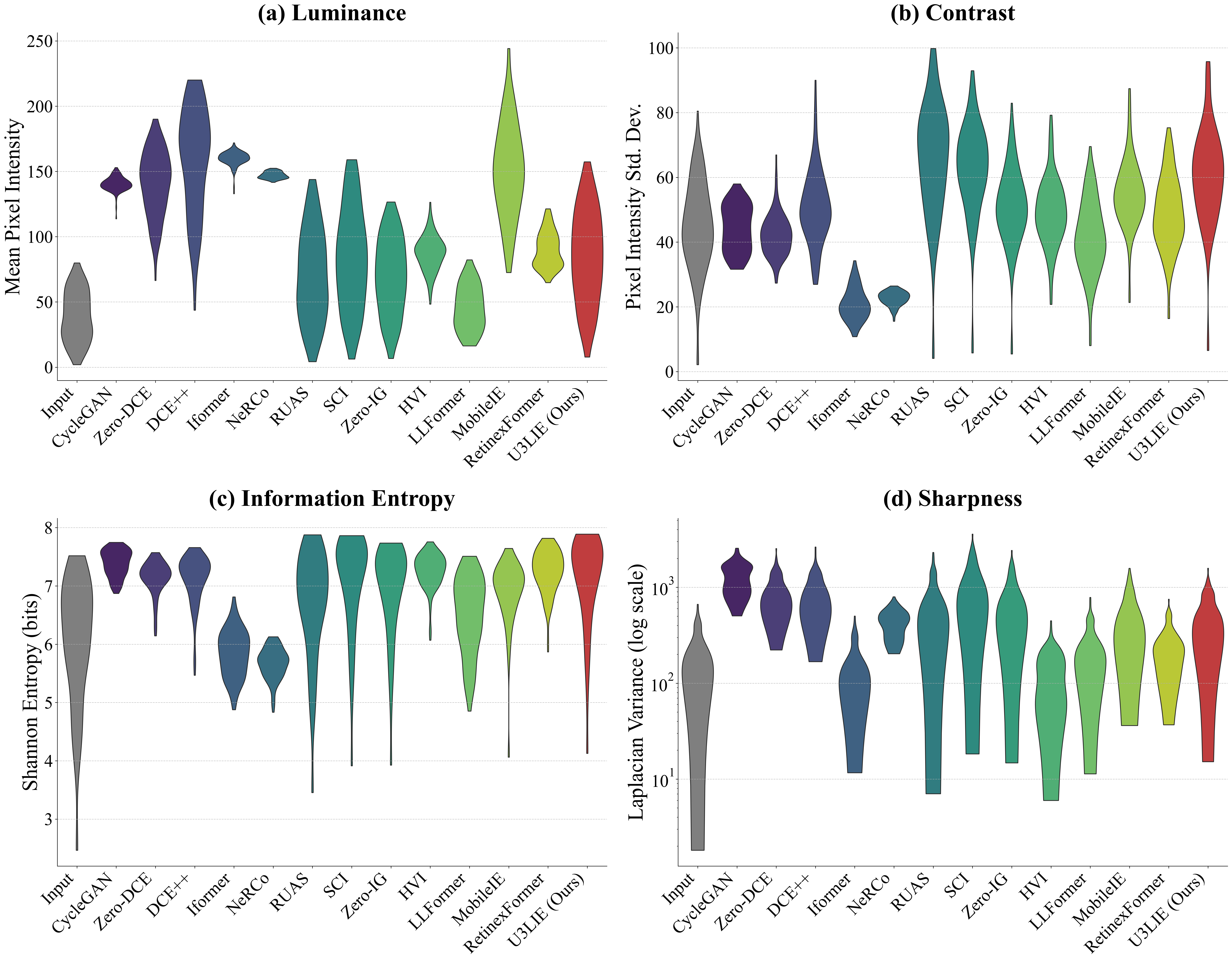}
	\caption{Quantitative analysis via distributional plots. We compare the results across four metrics: (a) Luminance, (b) Contrast, (c) Information Entropy, and (d) Sharpness.
	}
	\label{fig:distribution_comparison}
\end{figure}

\begin{figure*}[t]
	\centering
	\includegraphics[width=0.96\linewidth]{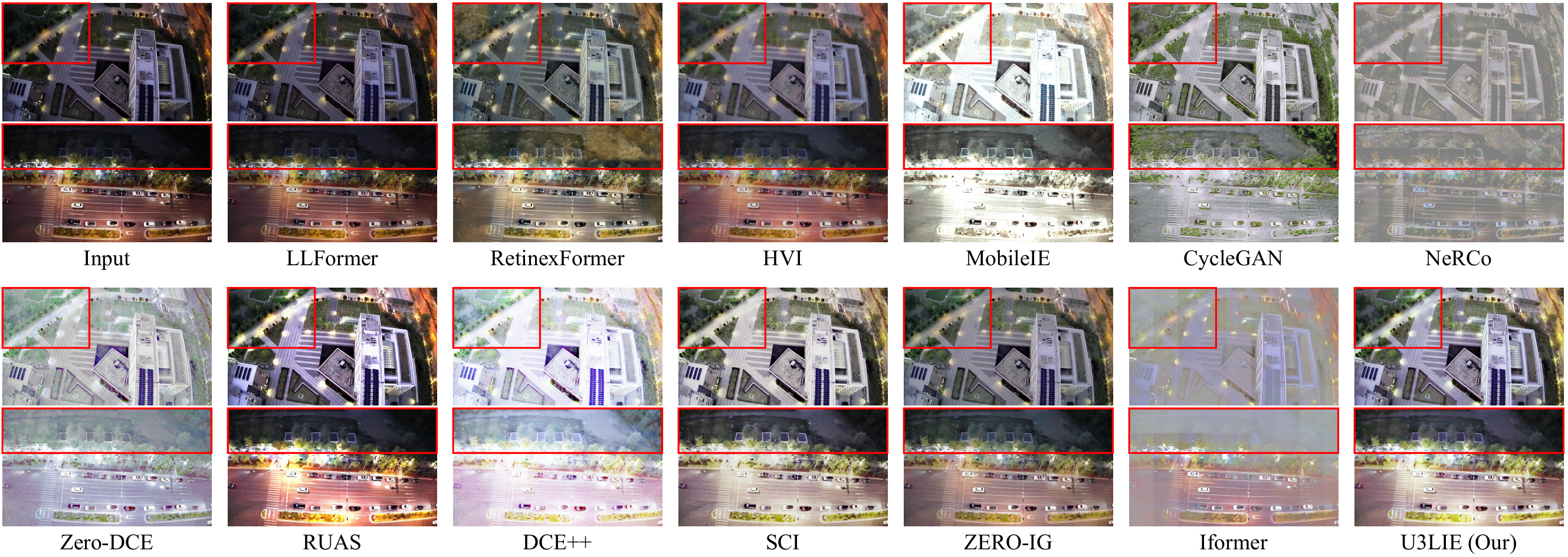} \vspace{-2mm}
	\caption{Visual comparison of our U3LIE with SOTA methods on the U3D dataset.}
	\label{fig:visual}\vspace{-2mm}
\end{figure*}

\subsection{Implementation Details}

\paragraph{Experimental Setup.}
All experiments were conducted on a single NVIDIA RTX 3090 GPU with 24GB of memory, using the PyTorch-based U3D toolkit on Ubuntu 20.04, and a fixed random seed of 2025 was set.

We used official implementations and default hyperparameters for all control methods. All models were evaluated on full-resolution 4K (3840$\times$2160) images from the U3D test set. For models that encountered Out-of-Memory (OOM) errors, we employed a patch-based evaluation strategy with a 256$\times$256 patch size and 64-pixel overlap, using Hann window blending for stitching. Validation was performed every 5 epochs, and the model with the best EEI score on the validation set was selected for the final test.

\paragraph{Training Configuration of U3LIE.}
Our model was trained for 100 epochs using the Adam optimizer with a weight decay of 10e$^{-4}$. We employed a 5-epoch linear learning rate warmup, then set the learning rate to 10e$^{-5}$, halving it every 50 epochs. We trained on 2048$\times$2048 patches with an effective batch size of 16, achieved through a batch size of 4 and 4 gradient accumulation steps. Gradient clipping was applied with a maximum norm of 0.05. For training, we used pairs of original images ($I_{in}$) and their pre-computed APA-processed counterparts ($I_{aug}$), generated once offline. The network receives $I_{aug}$ as input, while the loss function uses both the enhanced output $I_{enh}$ and the original input $I_{in}$. The loss weights in Eq.(\ref{eq:total_loss}) were set as follows: $\lambda_{tv}=100$, $\lambda_{spa}=4.0$, $\lambda_{col}=20.0$, and $\lambda_{int}=200$. For the $\mathcal{L}_{int}$ term itself, the local patch size was 16$\times$16, with targets for dark, bright, and global luminance set to 0.5, 0.6, and 0.6, respectively, and a global weight $\gamma_{global}=0.4$.

\subsection{Quantitative Results}
We present a comprehensive quantitative comparison against SOTA methods. 
For unsupervised evaluation, we include lightweight curve-based methods like Zero-DCE \cite{guo2020zeroreference}, RUAS \cite{liu2021retinexinspired}, DCE++ \cite{li2022learning}, SCI \cite{ma2022toward}, Zero-IG \cite{shi2024zeroig}, and Iformer \cite{li2025interpretable}. We also support unpaired methods such as CycleGAN \cite{zhu2017unpaired} and NeRCo \cite{yang2023implicit}. For comprehensive comparative analysis, we further integrate several supervised models, including LLFormer \cite{wang2023ultrahighdefinition},  RetinexFormer \cite{cai2023retinexformer}, HVI \cite{yan2025hvi}, and MobileIE \cite{yanmobileie}. For supervised baselines, we directly used their officially provided pre-trained weights for inference. 

Tab.~\ref{table:sota_comparison} shows that existing methods struggle with 4K aerial imagery. Many large, Transformer-based models (e.g., LLFormer, RetinexFormer) and some unpaired/unsupervised methods (NeRCo, Iformer) fail to process full-resolution images due to Out-of-Memory (OOM) errors on a 24GB GPU. Even when a patch-based strategy is employed (*), their inference speeds (FPS) are far too low for practical real-time applications on UAVs. While lightweight unsupervised methods such as Zero-DCE, RUAS, and DCE++ achieve real-time FPS, they exhibit poor perceptual quality on the U3D dataset, with PI scores all above 20. This indicates their struggle to handle the complex, non-uniform illumination of aerial scenes. Notably, unpaired methods like CycleGAN and NeRCo achieve better PI scores (5.76 and 10.25, respectively), but their large parameter counts and slow inference speeds result in extremely high (poor) EEI scores, making them unsuitable for edge deployment.

In contrast, U3LIE demonstrates a clear advantage across both efficiency and quality. 
On efficiency, U3LIE is exceptionally lightweight, requiring only \textbf{1.32K} parameters and \textbf{9.91G} FLOPs. This lightweight architecture enables a remarkable \textbf{23.80 FPS} on 4K images, over \textbf{3$\times$ faster} than the nearest real-time competitor (Zero-DCE at 7.76 FPS). 
On quality, U3LIE achieves a \textbf{PI score of 17.11}, significantly outperforming other lightweight methods and comparable to heavyweight GAN-based approaches without introducing their characteristic color artifacts.
This resulted in a final \textbf{EEI score of 16.03}, a new SOTA that is \textbf{3.5$\times$ better} than the prior best (DCE++, EEI=56.39). This outstanding result validates that our training strategy—the synergy of APA and $\mathcal{L}_{int}$—effectively unlocks the potential of a lightweight backbone for the demanding UHR-UAV context.

\paragraph{Distribution Analysis.}
Fig.~\ref{fig:distribution_comparison} presents a quantitative comparison of enhancement methods across four key image quality metrics. The distributions highlight the nuanced performance and trade-offs of each approach.

\textbf{Luminance (a):} Many methods, including DCE++ and MobileIE, consistently push luminance towards saturation, risking over-exposure and loss of detail in bright areas. In contrast, U3LIE demonstrates superior control, centering its luminance distribution in a perceptually balanced range while maintaining high consistency across the dataset.

\textbf{Contrast \& Entropy (b, c):} Crucially, this balanced illumination does not compromise detail restoration. U3LIE achieves top-tier performance in both contrast and information entropy. Its high and tight distributions in these metrics signify robust recovery of global structures and fine textures from dark regions, rivaling or exceeding all other methods.

\textbf{Sharpness (d):} The sharpness analysis reveals a trade-off. CycleGAN exhibits extreme sharpness, often at the cost of amplifying noise and introducing artifacts. U3LIE shows a more appropriate enhancement. It significantly improves clarity over the input while maintaining a controlled, moderate sharpness distribution, indicating effective detail enhancement without sacrificing image naturalness.

In summary, U3LIE excels at enhancing contrast and detail without introducing over-exposure or noise-related artifacts, demonstrating a superior trade-off.

\subsection{Qualitative Results}
Fig.~\ref{fig:visual} showcases U3LIE's superior performance on complex aerial scenes with non-uniform illumination. 

\textbf{Supervised methods} struggle with generalization to our U3D dataset. Lacking domain-specific training, they often under-enhance shadowed regions (LLFormer, RetinexFormer, and HVI) or cause over-exposure and color loss in bright areas (MobileIE).

\textbf{Unpaired methods} such as CycleGAN and NeRCo fail to maintain color fidelity. Their domain translation process introduces severe color artifacts, resulting in color distortion.

\textbf{Unsupervised methods} show mixed results. Lightweight models like Zero-DCE and DCE++ tend to over-expose light sources. RUAS suffers from both color distortion and under-enhancement, while Iformer leads to inconsistent illumination. Although SCI and Zero-IG show improvements, they still fail to match the balance of enhancement and efficiency achieved by our method.

\textbf{In contrast, U3LIE} excels by restoring visibility in dark areas without causing color distortion or over-exposing bright regions. As shown in the red boxes, our method produces results that are both visually realistic and rich in detail, faithfully preserving the original scene structure.

\begin{figure}[t]
	\centering
	\includegraphics[width=0.92\linewidth]{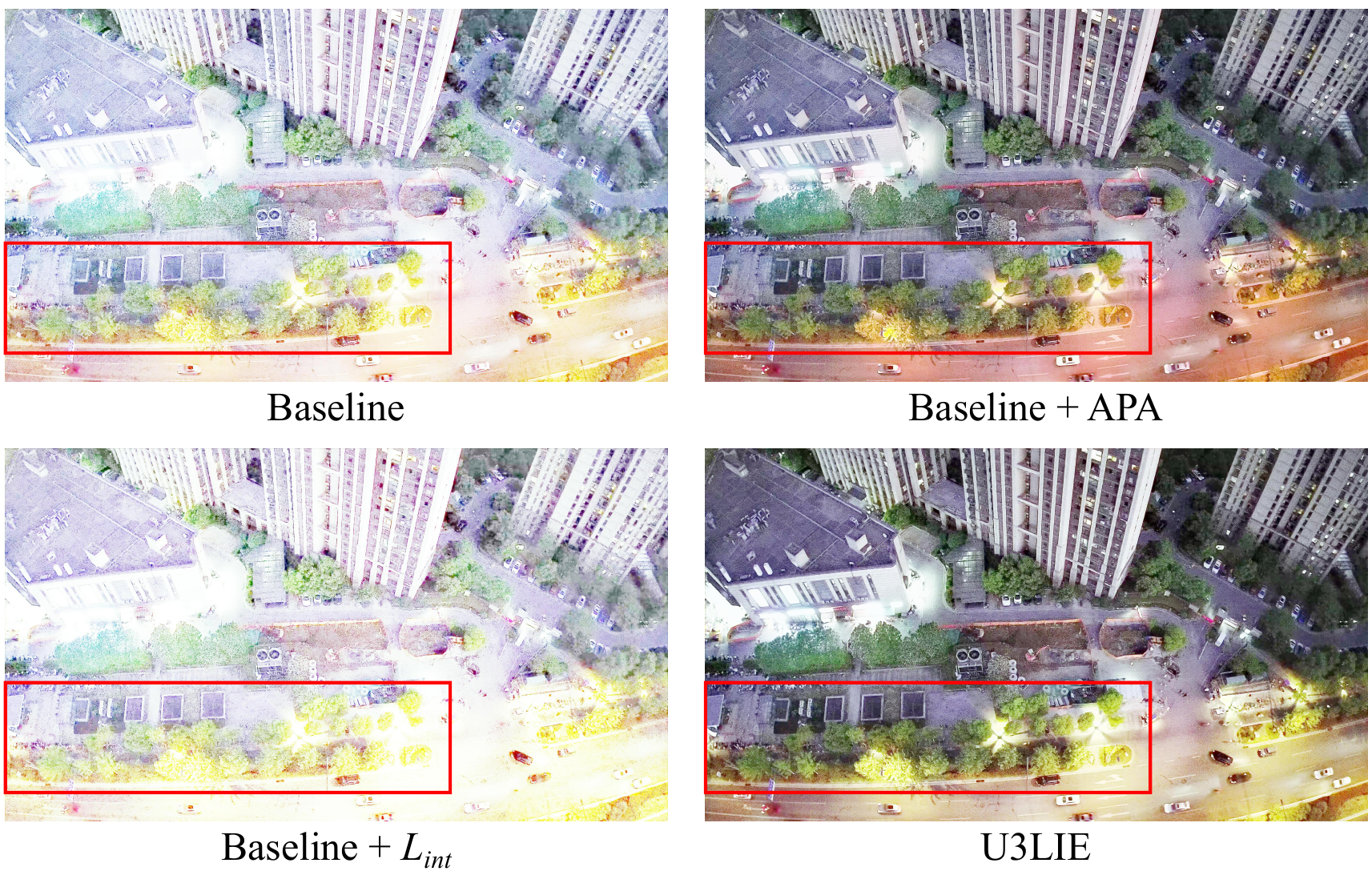}\vspace{-2mm}
	\caption{
		Ablation comparison of U3LIE on U3D test set.
	}
	\label{fig:abl}
\end{figure}

\begin{table}[t]\centering \small	\setlength\tabcolsep{6pt}	
	\renewcommand{\arraystretch}{1.05}
	\begin{tabular}{cc|ccc}\specialrule{0.8pt}{0pt}{0pt}
		\textbf{APA}	&\textbf{ $\mathcal{L}_{int}$} &{\textbf{NIQE $\downarrow$}}  	&\textbf{{BRISQUE}  $\downarrow$} 
		&\textbf{{PI}  $\downarrow$}  \\ \hline 
		&		 						&4.1987&38.0636&21.1311\\
		\ding{51} &		 				&4.5081&35.9074&20.2077\\
		&\ding{51} 						&4.2167&34.7033&19.4600\\ \hline
		\rowcolor[rgb]{0.9,0.9,0.9} 
		\multicolumn{2}{c}{U3LIE}\vline	&4.2954&29.9243&17.1098\\ \hline
	\end{tabular} \vspace{-2mm}
	\caption{Ablation study of U3LIE on U3D test set.} \label{Ablation_model}	
\end{table}

\subsection{Ablation Study}
\paragraph{U3LIE framework ablation.}
We conducted an ablation study to verify the effectiveness of our two key components: the APA module and the luminance-aware loss $\mathcal{L}{int}$. As shown in Tab.~\ref{Ablation_model}, the baseline DCE++ achieves a PI score of 21.13. Introducing APA alone improves the score to 20.21 by guiding the model to learn from a more consistent illumination distribution during training. Applying $\mathcal{L}{int}$ alone brings a larger improvement (PI = 19.46), as it directly encourages better luminance balance. When both components are used together in the full U3LIE framework, the PI score further drops to 17.11, demonstrating a clear synergy between APA and $\mathcal{L}{int}$. Fig.~\ref{fig:abl} supports these results visually: without APA, the model tends to over-enhance bright regions, while removing $\mathcal{L}{int}$ leads to color distortion. These findings confirm that both APA and $\mathcal{L}_{int}$ are crucial for achieving stable and natural enhancement.

\paragraph{Hyperparameter Analysis of $\mathcal{L}_{int}$.}
We ablate the key hyperparameters of the $\mathcal{L}_{int}$ and its associated weights, with results in Tab.~\ref{Ablation_hyp}. Our goal was to find a balance between perceptual quality (low PI) and visual fidelity.

First, we explored the target luminance intervals ($E_{dark}, E_{bright}$). We found that a narrow, mid-range interval like $[0.4, 0.6]$ outperforms wider or lower-biased ones (e.g., $[0.2, 0.8]$), confirming that constraining local brightness to a plausible mid-tone is effective.

Next, we analyzed the loss weights. The weight of our proposed loss, $\lambda_{int}$, is critical. Increasing $\lambda_{int}$ from 20 to 100 significantly improves the PI score (from 19.01 to 18.29). However, a high $\lambda_{int}$ combined with a low color loss weight ($\lambda_{col}=2$) leads to noticeable \textbf{color distortion} (Row 6). This occurs because the strong luminance guidance overwhelms the color constancy constraint, sacrificing color accuracy to meet brightness targets. This highlights a key trade-off between luminance correction and color fidelity.

Based on these findings, our final configuration ($\lambda_{tv}=100, \lambda_{spa}=4, \lambda_{col}=20, \lambda_{int}=200$ and an interval of $[0.5, 0.6]$) strikes an optimal balance. It achieves the best PI score while maintaining natural colors.

\begin{table}[t]\centering \small
	\sisetup{table-align-text-post=false} \setlength\tabcolsep{4.5pt}
	\renewcommand{\arraystretch}{1.05}
	\begin{tabular}{
			c c | 
			S[table-format=1.1]
			S[table-format=1.1]
			S[table-format=1.1]
			S[table-format=1.1] | 
			S[table-format=2.4]
		} \specialrule{0.8pt}{0pt}{0pt}
		\multicolumn{2}{c|}{\textbf{Loss Weights}} &
		\multicolumn{4}{c|}{\textbf{Hyperparameters of $\mathcal{L}_{int}$}} & 
		{\multirow{2}{*}{\textbf{PI $\downarrow$}}} \\
		$\lambda_{col}$ & $\lambda_{int}$ &{$E_{\text{dark}}$} 
		& {$E_{\text{bright}}$} & {$E_{\text{global}}$} & {$\gamma_{\text{global}}$} & \\ \hline 
		2		& 20 	& 0.2	& 0.8	& 0.6	& 0.4	& 19.3403 \\
		2		& 20 	& 0.3	& 0.7	& 0.6	& 0.6	& 19.0982 \\  
		2		& 10 	& 0.4	& 0.6	& 0.6	& 0.6	& 19.2415 \\
		2		& 20 	& 0.4	& 0.6	& 0.6	& 0.6	& 19.0064 \\
		2		& 100 	& 0.4	& 0.6	& 0.6	& 0.6	& 18.2944 \\
		2		& 200 	& 0.4	& 0.6	& 0.6	& 0.6	& 16.9 {(CD)} \\
		50		& 500 	& 0.5	& 0.6	& 0.6	& 0.4	& 17.1003 \\ \hline
		\rowcolor[rgb]{0.9,0.9,0.9} 20		& 200 	& 0.5	& 0.6	& 0.6	& 0.4	& 17.1098 \\ \hline
	\end{tabular} \vspace{-2mm}
	\caption{Ablation study of $\mathcal{L}_{int}$ and its hyperparameters. In all experiments, the loss weights $\lambda_{tv}$ and $\lambda_{spa}$ are fixed to 100 and 4, respectively. CD denotes color distortion.}
	\label{Ablation_hyp}
\end{table}

\section{Conclusion and Future Work}	\label{conclusion}
This paper addresses the challenges of unsupervised, UHR LIE for UAVs by introducing a foundational suite of tools: the \textbf{U3D} and \textbf{U3D toolkit}, the deployment-aware \textbf{EEI} metric, and the highly efficient \textbf{U3LIE} framework. U3LIE achieves a SOTA balance of visual quality and real-time performance, providing a robust baseline for developing practical solutions. For future work, we will quantify the benefit to downstream tasks and explore joint optimization to maximize performance in safety-critical UAV applications.

\setcounter{page}{1}
\appendix
\onecolumn

\section*{Appendix: Supplementary Materials for ``Unsupervised Ultra-High-Resolution UAV Low-Light Image Enhancement: A Benchmark, Metric and Framework''} \label{Appendix}

This supplementary document provides additional implementation and analysis details that complement the main paper. It aims to facilitate reproducibility and deepen understanding of our proposed U3D benchmark, EEI metric, and U3LIE framework.

The appendix is organized as follows:
\begin{itemize}
	\item \textbf{Appendix A} provides a comprehensive analysis and comparison of low-light datasets, including details on the U3D data collection protocol, further justifying its necessity and novelty.
	\item \textbf{Appendix B} presents a detailed breakdown and robustness analysis of the EEI metric.  
	\item \textbf{Appendix C} details the architectural design of the APA. 
	\item \textbf{Appendix D} includes a discussion of limitations and failure cases.
	\item \textbf{Appendix E} provides additional visual results and qualitative comparisons.
\end{itemize}

\section{Dataset Comparison and Analysis}

The development of robust Low-Light Image Enhancement (LIE) algorithms is critically dependent on the availability of high-quality, relevant benchmark datasets. While numerous datasets have been proposed, they were largely designed for ground-level computer vision tasks and fail to address the unique constellation of challenges posed by Unmanned Aerial Vehicle (UAV) imaging. This appendix provides an in-depth comparative analysis to rigorously position our U3D dataset within the existing landscape and to articulate its foundational importance for the future of 24/7 autonomous aerial systems.

Table \ref{tab:detailed_dataset_comparison} presents a detailed, multi-faceted comparison, which is followed by a narrative analysis structured around the key gaps that U3D is designed to fill.

\begin{table*}[!ht]	\centering
	\renewcommand{\arraystretch}{1.3}
	\resizebox{\linewidth}{!}{%
		\begin{tabular}{@{}llccllllll@{}}	\toprule
			\textbf{Dataset} & \textbf{Reference} & \textbf{Resolution} & \textbf{Type} & \textbf{Perspective} & \textbf{Scene Characteristics} & \textbf{Primary Task} & \textbf{Annotations} & \textbf{Size (Low)} & \textbf{Key Limitations for UAV LIE} \\
			\midrule
			\textbf{U3D (Ours)} & \textbf{This Work} & \textbf{3840$\times$2160} & \textbf{Unpaired} & \textbf{Aerial (Top-down)} & \textbf{Urban/Suburban, Non-uniform light} & \textbf{Enhancement} & \textbf{None} & \textbf{1,000} & \textbf{Specifically designed for this task} \\
			\midrule
			\multicolumn{10}{l}{\textit{\textbf{Supervised Ground-Level Datasets}}} \\
			\midrule
			LOL-v1 & \cite{wei2018deep}  & 400$\times$600 & Paired & Ground-level & Indoor/Outdoor, Mixed lighting & Enhancement & None & 500 & Low-res; Paired data; Ground-level view. \\
			LOL-v2 & \cite{wei2018deep} & 400$\times$600 & Paired & Ground-level & Real \& Synthetic scenes & Enhancement & None & 789 & Same limitations as LOL-v1, still low-resolution. \\
			SID & \cite{chen2018learning}  & 4240$\times$2832 & Paired (RAW) & Ground-level & Indoor/Outdoor, Extreme darkness & Denoising/Enhance. & None & 5,094 & Focus on RAW data; Paired data; Ground-level. \\
			UHD-LL & \cite{wang2023ultrahighdefinition} & 3840$\times$2160 & Paired & Ground-level & Street/Nightscapes, Non-uniform light & Enhancement & None & 8,099 & Paired data; Ground-level perspective. \\
			FiveK & \cite{bychkovsky2011learning}  & $\sim$5K & Semi-Paired & Ground-level & General scenes, Variable light & Retouching & None & 5,000 & Not specific to low-light; Paired data is synthetic. \\
			\midrule
			\multicolumn{10}{l}{\textit{\textbf{Unsupervised Ground-Level Datasets}}} \\
			\midrule
			ExDark & \cite{lohia2019exdark}  & Variable (Low) & Unpaired & Ground-level & Object-centric, 12 categories & Enhancement & B-Boxes & 7,363 & Low/inconsistent resolution; Lacks complex scenic context. \\
			NPE & \cite{wang2013naturalness}  & Variable (Low) & Unpaired & Ground-level & Mixed scenes, Natural images & Enhancement & None & 85 & Very small size; Low resolution; Outdated. \\
			DICM & \cite{lee2013contrast}  & 512$\times$384 & Unpaired & Ground-level & General photography & Enhancement & None & 69 & Very low resolution; Too small for deep learning. \\
			LIME & \cite{guo2017lime} & Variable (Low) & Unpaired & Ground-level & Mixed scenes for qualitative eval & Enhancement & None & 10 & A qualitative set, not for training/testing. \\
			\midrule
			\multicolumn{10}{l}{\textit{\textbf{Driving-Specific Datasets (Forward-Facing Perspective)}}} \\
			\midrule
			Dark Zurich & \cite{sakaridis2019guided} & 1920$\times$1080 & Unpaired & Driving & Urban/Suburban driving at dusk/night & Segmentation & Seg. Masks & 8,779 & Perspective mismatch; Not top-down; Moderate resolution. \\
			BDD100K-Night & \cite{yu2020bdd100k}  & 1280$\times$720 & Unpaired & Driving & Highways, City streets at night & Detection/Seg. & B-Boxes, etc. & $\sim$10,000 & Perspective mismatch; Lower resolution (720p). \\
			NightCity & \cite{9591338}  & 1920$\times$1080 & Unpaired & Driving & Urban driving across cities & Segmentation & Seg. Masks & 4,297 & Perspective mismatch; Focus on high-level tasks. \\
			\midrule
			\multicolumn{10}{l}{\textit{\textbf{Aerial Datasets (Not for LIE)}}} \\
			\midrule
			VisDrone & \cite{zhu2021vision} & $\sim$1920$\times$1080 & N/A & Aerial (UAV) & Mostly daytime scenes & Detection/Tracking & B-Boxes & N/A & Lacks low-light focus; Not for enhancement task. \\
			UAVDT & \cite{du2018the} & $\sim$1080p & N/A & Aerial (UAV) & Mostly daytime/dusk traffic scenes & Tracking & B-Boxes & N/A & Lacks sufficient night scenes for enhancement. \\
			AU-AIR & \cite{bozcan2020au}  & 1920$\times$1080 & N/A & Aerial (UAV) & Daytime, Various weather & Detection & B-Boxes & N/A & No low-light data. \\
			\bottomrule
		\end{tabular}%
	}
	\caption{A detailed and comprehensive comparison of low-light and aerial datasets. U3D is the first to uniquely combine an aerial perspective, ultra-high resolution, an unpaired data structure, and scenes with extreme non-uniform illumination, making it the essential benchmark for real-world UAV LIE.}	\label{tab:detailed_dataset_comparison}
\end{table*}

\subsection{The Critical Gaps in Existing Benchmarks}
Our analysis reveals four fundamental gaps in the existing dataset landscape that have hindered the development of practical LIE models for UAVs. U3D is the first benchmark to systematically address all four.

\paragraph{The Resolution and Detail Gap.}
UAVs fly at significant altitudes, causing objects of interest on the ground (e.g., people, vehicles, debris) to occupy a very small portion of the image. Consequently, \textbf{Ultra-High Resolution (UHR)} is not a luxury but a necessity for any meaningful surveillance or inspection task. The vast majority of established LIE datasets, such as LOL (400$\times$600), ExDark (variable low-res), and BDD100K (1280$\times$720), lack the resolution required to retain these critical details. While UHD-LL provides 4K data, its ground-level perspective and supervised nature render it unsuitable. U3D's provision of 4K aerial imagery directly confronts this gap, enabling the development and evaluation of models that can enhance fine-scale details vital for real-world UAV missions. This UHR constraint also makes computational efficiency, measured by our EEI metric, a first-class citizen in model design.

\paragraph{The Paired-Data Impasse.}
Supervised learning, powered by datasets like LOL, SID, and UHD-LL, has been a dominant paradigm in image enhancement. However, it relies on a critical assumption: the availability of perfectly aligned low-light and ground-truth normal-light image pairs. For a dynamic, constantly moving UAV, this is a practical impossibility. Factors such as minute changes in flight path, atmospheric conditions, seasonal variations, and alterations in the ground environment between day and night make pixel-perfect alignment infeasible. Therefore, any robust, scalable UAV enhancement solution must be \textbf{unsupervised}. U3D is constructed from the ground up as an unpaired dataset, directly reflecting this real-world constraint and fostering the development of unsupervised frameworks like U3LIE that do not depend on unattainable ground-truth pairs.

\paragraph{The Perspective Mismatch.}
The world looks profoundly different from above. Datasets captured from a ground-level (e.g., LOL, ExDark) or a forward-facing driving perspective (e.g., Dark Zurich, BDD100K) exhibit different object scales, shadow patterns, and light source characteristics. A top-down aerial view, as captured in U3D, features vast, contiguous dark areas (e.g., rooftops, fields, parks) punctuated by intense, point-like light sources (streetlights, headlights). This unique \textbf{aerial perspective} creates a distinct image domain. Models trained on ground-level or driving data struggle to generalize to this domain, often failing to handle the unique spatial distribution of light and shadow. U3D provides the first domain-specific benchmark to train and validate models specifically for the challenges of aerial vision.

\paragraph{The Illumination Complexity Gap.}
Nighttime aerial scenes exhibit a form of \textbf{severe, non-uniform illumination} that is more challenging than in typical ground-level imagery. The dynamic range is extreme, with deep, information-starved shadows coexisting with saturated highlights and specular blooming artifacts in a single frame. This is not merely a few isolated bright regions in a dark image; it's about a complex interplay of varied light intensities across a massive field of view. Existing datasets often feature more diffuse or less extreme lighting conditions. The challenging scenes in U3D expose the limitations of conventional loss functions and enhancement techniques, which tend to either amplify noise in dark regions or crush details in bright ones. This complexity necessitates novel approaches, such as our Luminance Interval Loss ($\mathcal{L}_{int}$), which is specifically designed to manage this trade-off and achieve a balanced enhancement across the entire high-dynamic-range scene.

\subsection{U3D: A Foundational Benchmark for Robust UAV Vision}

\begin{figure*}[h!]	\centering
	\includegraphics[width=\linewidth]{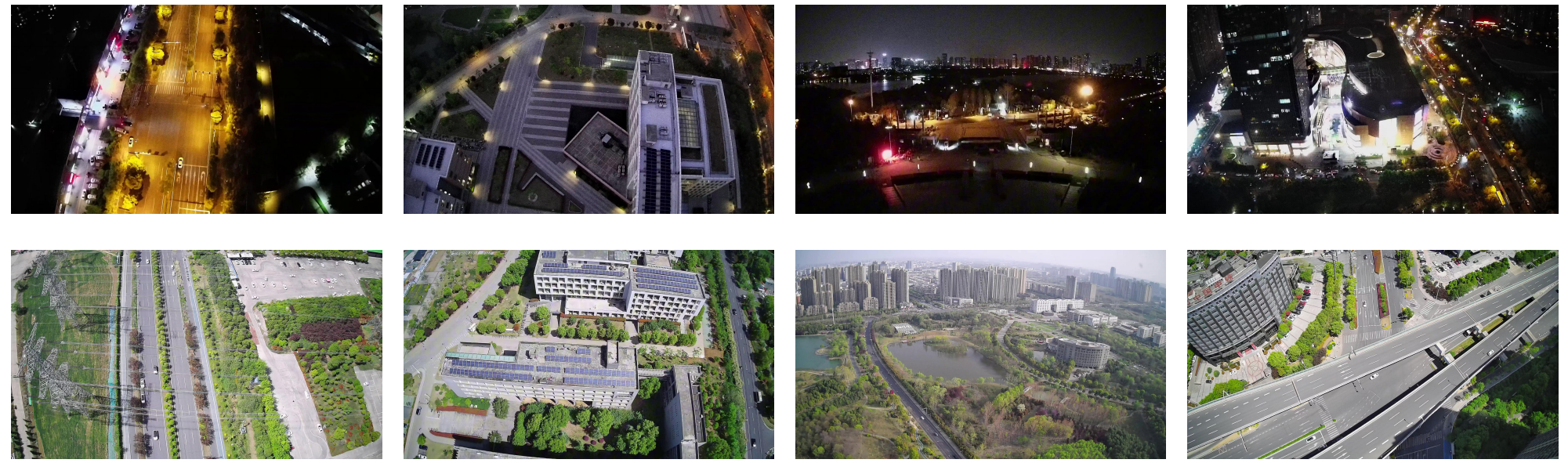} 
	\caption{Visual examples from our U3D dataset. The top row showcases UHR low-light images captured by UAVs across diverse scenes. The bottom row presents non-paired daylight images from similar environments, highlighting the dataset's structure for unsupervised learning. The variety of scenes poses significant challenges in terms of non-uniform illumination and detail preservation.}
	\label{fig:comp}
\end{figure*}

To address the critical gaps in existing benchmarks, we introduce U3D, the first dataset specifically designed for unsupervised, UHR LIE on Unmanned Aerial Vehicles (UAVs).

\textbf{Dataset Construction and Characteristics.} The U3D benchmark comprises \textbf{1,000 UHR (3840$\times$2160) low-light images and 1,000 non-paired daytime counterparts.} All images were captured using a professional-grade UAV equipped with a 3-axis gimbal to ensure high quality and minimize motion blur. To ensure high diversity, flights were conducted over a diverse array of real-world environments with altitudes ranging from 30m to 120m and camera angles spanning from forward-looking to nadir (top-down) views. As illustrated in Figure~\ref{fig:comp}, the dataset covers a wide array of scenes, including:
\begin{itemize}
	\item \textbf{Commercial \& Business Hubs:} High-density Central Business Districts (CBDs), shopping malls, and brightly lit commercial avenues.
	\item \textbf{Residential Areas:} A mix of dense urban apartment complexes and sprawling suburban neighborhoods.
	\item \textbf{Public \& Institutional Zones:} University campuses, schools, public parks, and tourist attractions.
	\item \textbf{Transportation Infrastructure:} Major arterial roads, city streets, and complex interchanges.
\end{itemize}
Critically, to capture authentic sensor characteristics, all images were recorded in automatic mode, allowing the on-board system to dynamically adjust ISO (typically 800-6400) and shutter speed. No in-camera noise reduction or post-processing was applied, providing direct sRGB outputs in `.jpg' format. This preserves the genuine sensor noise and compression artifacts that real-world models must learn to handle.

\textbf{Significance and Impact.} U3D represents a constructed benchmark that advances the state of the art in UAV-based low-light vision research. Unlike conventional datasets, our benchmark systematically addresses four key challenges in real-world aerial imaging: (1) ultra-high resolution processing, (2) unpaired data learning, (3) perspective variations, and (4) illumination heterogeneity. This provides the first comprehensive framework for developing and evaluating LIE algorithms specifically optimized for UAV platforms. Consequently, our work transitions the research paradigm from limited ground-level scenarios to practical aerial vision systems capable of reliable around-the-clock operation.

\section{Edge Efficiency Index (EEI) Details}
This section presents a rigorous, multi-faceted efficiency analysis centered on the proposed Edge Efficiency Index (EEI), a composite metric designed to provide a holistic and hardware-agnostic assessment of on-device model performance. We first define the EEI framework and its components. We then apply it to the \texttt{DCE++ light} model, evaluating its performance scaling on a single high-end GPU. Finally, we validate the robustness of the EEI metric itself through a cross-platform comparison involving desktop and edge devices.

\subsection{The EEI Framework: Definition and Calculation}
Our evaluation is based on the Edge Efficiency Index (EEI), defined as:
$\text{EEI} = \text{PI} \times (w_T \cdot \text{TF} + w_C \cdot \text{CF} + w_R \cdot \text{RF})$, 
where PI is the Perceptual Index score, reflecting task-specific performance. To focus on hardware efficiency for this analysis, we normalize the PI score to a constant value of 100. The core of the metric is the Efficiency Factor, a weighted sum of three sub-metrics:
\begin{itemize}
	\item \textbf{Time Factor (TF):} Normalized inference latency, reflecting temporal efficiency.
	\item \textbf{Complexity Factor (CF):} Normalized static model complexity (FLOPs and parameters), reflecting computational cost.
	\item \textbf{Resource Factor (RF):} Normalized peak memory footprint (VRAM/RAM), reflecting runtime resource consumption.
\end{itemize}
The weights $w_T, w_C, w_R$ (summing to 10) represent the relative importance assigned to each factor. We propose the \textbf{8:1:1 ($w_T=8, w_C=1, w_R=1$) weighting as a pragmatic default} for evaluating on-device models. This policy allocates 80\% of the importance to latency (TF), reflecting the paramount importance of user-perceived speed. The remaining 20\% is divided between model complexity (CF) and memory footprint (RF), acknowledging them as critical but secondary constraints. This weighting scheme emphasizes latency (80\%) to reflect its primary impact on user experience and real-time constraints, while accounting for model complexity and memory usage as secondary but important deployment factors.

\subsubsection{Detailed Factor Calculation}
The following section provides the precise mathematical formulation for each factor, as implemented in our evaluation script.

\paragraph{Baseline Definition}
All factors are normalized against a hardware-specific baseline to ensure fair, cross-device comparisons. This baseline is established by running a standard MobileNetV2 model on a 4K resolution input ($3840 \times 2160$ pixels) on the target hardware. The following reference values are obtained:
\begin{itemize}
	\item $\text{Time}_{\text{ref}}$: The average inference time (in seconds) of MobileNetV2 at 4K.
	\item $\text{FLOPs}_{\text{ref}}$: The floating-point operations of MobileNetV2 for a 4K input.
	\item $\text{Params}_{\text{ref}}$: The number of trainable parameters in MobileNetV2.
	\item $\text{Mem}_{\text{ref}}$: The peak memory consumption (VRAM on GPU, RSS increment on CPU) of MobileNetV2 at 4K.
\end{itemize}

\paragraph{Time Factor (TF)}
The Time Factor measures the model's temporal efficiency relative to the hardware's baseline performance, normalized for resolution. It quantifies how fast the model processes each pixel compared to the baseline model. A lower TF is better.
\[
\text{TF} = \frac{\text{Time}_{\text{model}}}{\text{Time}_{\text{ref}}} \times \frac{\text{Pixels}_{\text{ref}}}{\text{Pixels}_{\text{model}}}
\]
where $\text{Time}_{\text{model}}$ is the measured inference time of the evaluated model, and $\text{Pixels}$ denotes the total pixel count for the respective resolution.

\paragraph{Complexity Factor (CF)}
The Complexity Factor measures the static, hardware-independent complexity of the model. A lower CF is better. It is a weighted average of the normalized FLOPs and parameters:
\[
\text{CF} = 0.5 \times \frac{\text{FLOPs}_{\text{model}}}{\text{FLOPs}_{\text{ref}}} + 0.5 \times \frac{\text{Params}_{\text{model}}}{\text{Params}_{\text{ref}}}
\]

\paragraph{Resource Factor (RF)}
The Resource Factor measures the model's runtime memory consumption relative to the baseline. A lower RF is better. It is a direct ratio of the peak memory usage:
\[
\text{RF} = \frac{\text{Mem}_{\text{model}}}{\text{Mem}_{\text{ref}}}
\]
where $\text{Mem}_{\text{model}}$ is the peak memory usage of the model at its \textbf{specific test resolution}, while $\text{Mem}_{\text{ref}}$ is from the baseline at a \textbf{fixed 4K resolution}. This intentionally penalizes models whose memory usage scales poorly with input size.

\subsection{Algorithmic Implementation}

\begin{algorithm}[h!]
	\caption{EEI (Edge Efficiency Index) Calculation Pipeline}
	\label{alg:eei_pipeline}
	\begin{algorithmic}[1]
		\Require Model $M$, Test Image Size $S_{test}$, Device $D$
		\Ensure EEI score
		
		\Statex
		\Function{Get-Hardware-Baseline}{$D$}
		\State $M_{base} \gets \text{MobileNetV2 on device } D$
		\State $S_{base} \gets (1, 3, 3840, 2160)$ \Comment{Fixed 4K baseline size}
		\State $Input_{base} \gets \text{Random tensor of size } S_{base}$
		\State \textbf{try:}
		\State \hspace{\algorithmicindent} $(FLOPs_{base}, Params_{base}) \gets \Call{profile}{M_{base}, Input_{base}}$
		\State \textbf{except} Out-of-Memory (OOM):
		\State \hspace{\algorithmicindent} $(FLOPs_{base}, Params_{base}) \gets (\text{None}, \text{None})$
		\State $Mem_{base} \gets \Call{measure\_peak\_memory}{M_{base}, Input_{base}}$
		\State $Time_{ref} \gets \Call{measure\_avg\_inference\_time}{M_{base}, Input_{base}}$
		\State \textbf{return} ($Time_{ref}$, $FLOPs_{base}$, $Params_{base}$, $Mem_{base}$)
		\EndFunction
		
		\Statex
		\Function{Get-Model-Performance}{$M, S_{test}, D$}
		\State $Input_{test} \gets \text{Random tensor of size } S_{test}$
		\State \textbf{try:}
		\State \hspace{\algorithmicindent} $(FLOPs_{model}, Params_{model}) \gets \Call{profile}{M, Input_{test}}$
		\State \textbf{except} OOM:
		\State \hspace{\algorithmicindent} $(FLOPs_{model}, Params_{model}) \gets (\text{None}, \text{None})$
		\State \textbf{try:}
		\State \hspace{\algorithmicindent} $Time_{model} \gets \Call{measure\_avg\_inference\_time}{M, Input_{test}}$
		\State \hspace{\algorithmicindent} $Mem_{model} \gets \Call{measure\_peak\_memory}{M, Input_{test}}$
		\State \textbf{except} OOM: \Comment{Fallback to patch-based processing}
		\State \hspace{\algorithmicindent} $(\_, Time_{model}, Mem_{model}) \gets \Call{process\_image\_robustly}{M, Input_{test}}$
		\State \textbf{return} ($Time_{model}$, $FLOPs_{model}$, $Params_{model}$, $Mem_{model}$)
		\EndFunction
		
		\Statex \Comment{--- Main Calculation ---}
		\State $(Time_{ref}, FLOPs_{base}, \dots) \gets \Call{Get-Hardware-Baseline}{D}$
		\State $(Time_{model}, FLOPs_{model}, \dots) \gets \Call{Get-Model-Performance}{M, S_{test}, D}$
		\State $PI \gets \Call{Calculate-Perceptual-Index}{M, S_{test}}$ \Comment{Using NIQE \& BRISQUE}
		\Statex
		\State $Pixels_{test} \gets \prod S_{test}$; \ $Pixels_{base} \gets \prod S_{base}$
		\State $TF \gets (Time_{model} / Time_{ref}) \times (Pixels_{base} / Pixels_{test})$ \Comment{Time Factor}
		\State $RF \gets Mem_{model} / Mem_{base}$ \Comment{Resource Factor}
		\Statex
		\If{$FLOPs_{model}$ \textbf{and} $Params_{model}$ are available}
		\State $CF \gets 0.5 \times (FLOPs_{model} / FLOPs_{base}) + 0.5 \times (Params_{model} / Params_{base})$ \Comment{Complexity Factor}
		\State $E_{norm} \gets 0.8 \cdot TF + 0.1 \cdot CF + 0.1 \cdot RF$
		\Else \Comment{Fallback for OOM during profiling}
		\State $E_{norm} \gets 0.9 \cdot TF + 0.1 \cdot RF$
		\EndIf
		\Statex
		\State $EEI \gets PI \times E_{norm}$
		\State \textbf{return} $EEI$
	\end{algorithmic}
\end{algorithm}

To ensure clarity and facilitate adoption by the community, we formalize the entire EEI calculation process in Algorithm~\ref{alg:eei_pipeline}. The pipeline is designed to be robust and hardware-agnostic. It begins by establishing a hardware-specific baseline using a standard MobileNetV2 model at a fixed 4K resolution (Lines 1-11). Then, it measures the performance of the model under evaluation ($M$) at its specific test resolution ($S_{test}$) (Lines 12-23).

A key feature of our implementation is its robustness to Out-of-Memory (OOM) errors, which are common when evaluating large models on UHR inputs. If profiling or full-size inference fails due to OOM, the algorithm gracefully degrades: it can fall back to a patch-based processing mode for timing measurements (Line 22) and automatically switches to a complexity-agnostic weighting scheme for the final EEI calculation (Line 35). This ensures that all models can be fairly evaluated, even those that cannot run directly on high-resolution inputs.

\subsection{Experimental Analysis of \texttt{DCE++ light}}
We now apply the EEI framework to evaluate the \texttt{DCE++ light} model. The analysis is two-fold: first, we examine its performance scalability and the impact of different weighting policies on a single platform; second, we assess the consistency of the EEI metric across diverse hardware.

\subsubsection{Performance Scaling and Weighting Ablation on a Single Platform}
The initial evaluation was conducted on an NVIDIA GeForce RTX 3090 GPU across six resolutions, from 480p to 8K. The results, summarized in Table~\ref{tab:eei_horizontal}, reveal the model's efficiency profile under varying computational loads.

\sisetup{
	round-mode=places,
	round-precision=2,
	table-format=3.2 
}

\begin{table}[h!]
	\centering
	\small
	\setlength{\tabcolsep}{5.6pt} 
	\renewcommand{\arraystretch}{1.3}
	\begin{tabular}{c
			S[table-format=3.2] S[table-format=3.2]
			S[table-format=2.2] S[table-format=2.4] 
			S[table-format=1.3] S[table-format=1.3] S[table-format=1.3] |
			S[table-format=3.2] S[table-format=3.2] S[table-format=3.2] 
			S[table-format=3.2] S[table-format=3.2]} 
		\toprule
		\multirow{2.5}{*}{\textbf{Resolution}} &
		{\multirow{2.5}{*}{\textbf{\makecell[c]{Time \\ (ms)}}}} &
		{\multirow{2.5}{*}{\textbf{FPS}}} &
		{\multirow{2.5}{*}{\textbf{\makecell[c]{FLOPs \\ (G)}}}} &
		{\multirow{2.5}{*}{\textbf{\makecell[c]{Mem \\ (GB)}}}} &
		{\multirow{2.5}{*}{\textbf{TF}}} &
		{\multirow{2.5}{*}{\textbf{CF}}} &
		{\multirow{2.5}{*}{\textbf{RF}}} &
		\multicolumn{5}{c}{\textbf{EEI Score for Weight Scenario}} \\
		\cmidrule(lr){9-13}
		& & & & & & & & {\textbf{8:1:1}} & {\textbf{9:0.5:0.5}} & {\textbf{6:2:2}} & {\textbf{9:0:1}} & {\textbf{10:0:0}} \\
		\midrule
		854$\times$480   & 2.07   & 482.32 & 0.49  & 0.178  & 1.033 & 0.005 & 0.075  & 83.45  & 93.38  & 63.58  & 93.73  & 103.31 \\
		1280$\times$720  & 4.69   & 213.19 & 1.10  & 0.359  & 1.040 & 0.010 & 0.152  & 84.79  & 94.38  & 65.62  & 95.08  & 103.96 \\
		1920$\times$1080 & 10.48  & 95.45  & 2.48  & 0.719  & 1.032 & 0.023 & 0.304  & 85.83  & 94.51  & 68.45  & 95.92  & 103.20 \\
		2560$\times$1440 & 18.47  & 54.14  & 4.41  & 1.264  & 1.024 & 0.041 & 0.534  & 87.63  & 94.99  & 72.91  & 97.45  & 102.35 \\
		3840$\times$2160 & 41.82  & 23.91  & 9.91  & 2.756  & 1.030 & 0.092 & 1.164  & 94.96  & 98.98  & 86.92  & 104.34 & 102.99 \\
		7680$\times$4320 & 167.39 & 5.97   & 39.65 & 10.779 & 1.031 & 0.367 & 4.554  & 131.65 & 117.35 & 160.25 & 138.28 & 103.05 \\
		\bottomrule
	\end{tabular}
	\caption{EEI evaluation of the \texttt{DCE++ light} model (PI Score = 100) on an NVIDIA GeForce RTX 3090 GPU. Time Factor ($TF$), Complexity Factor ($CF$), and Resource Factor ($RF$) are shown. EEI scores are presented for different weight scenarios: Default (8:1:1), 9:0.5:0.5, 6:2:2, No CF (9:0:1), and Time Only (10:0:0).}
	\label{tab:eei_horizontal}
\end{table}

The physical metrics (Time, FLOPs, Memory) scale predictably with the input pixel count. An 81-fold increase in pixels from 480p to 8K results in a commensurate \textasciitilde81-fold increase in latency and FLOPs, confirming the model's linear computational scalability. However, peak memory usage grows substantially, reaching \SI{10.78}{\giga\byte} at 8K, which may be a constraint for memory-limited devices. The normalized factors provide deeper insights. The \textbf{Time Factor (TF)} remains stable around 1.03 across all resolutions, indicating consistent per-pixel processing efficiency relative to the baseline. The \textbf{Complexity Factor (CF)} stays well below 1.0, confirming the model's lightweight design. In stark contrast, the \textbf{Resource Factor (RF)} escalates from 0.075 to 4.554, identifying memory consumption as the primary efficiency bottleneck at ultra-high resolutions, as it scales significantly faster than the baseline.

The ablation study on EEI weighting policies further illuminates this profile. This study is not intended to find a single ``best'' score, but rather to demonstrate how the perceived efficiency of a model changes based on application-specific priorities.
\begin{itemize}
	\item \textbf{6:2:2.} This policy gives greater weight to complexity and resources. Consequently, \texttt{DCE++ light} achieves its lowest EEI scores under this policy for resolutions up to 4K, as the policy rewards the model's primary strengths (low CF and RF). However, this trend reverses at 8K, where the extreme RF value (4.554) is heavily penalized, resulting in the worst EEI score (160.25). This inversion powerfully illustrates that a model optimal for one set of constraints can be suboptimal for another.
	\item \textbf{``Latency-Priority'' (9:0.5:0.5).} By further emphasizing speed, this policy highlights the model's stable but not superior TF (\textasciitilde1.03). The scores are consistently higher than the Default policy at lower resolutions but become more favorable at 8K, where the impact of the high RF is minimized.
	\item \textbf{``Time-Only'' (10:0:0).} In this boundary case, the EEI score is governed solely by TF. The stable score of \textasciitilde103 across all resolutions isolates the model's temporal performance, confirming its resolution-agnostic per-pixel efficiency if all other constraints were ignored.
\end{itemize}

\subsubsection{Cross-Platform Consistency of the EEI Metric}
To validate the robustness of the EEI framework, we extended the evaluation to two additional platforms: a modern desktop NVIDIA RTX 4060 GPU and an edge-computing NVIDIA Jetson Xavier device. The results are presented in Table~\ref{tab:eei_combined_gpus}.

\sisetup{
	round-mode=places,
	round-precision=2,
	table-format=3.2 
}

\begin{table}[h!]
	\centering
	\small
	\setlength{\tabcolsep}{4pt} 
	\renewcommand{\arraystretch}{1.3}
	\begin{tabular}{cc
			S[table-format=3.2] S[table-format=3.2]
			S[table-format=2.2] S[table-format=2.4] 
			S[table-format=1.3] S[table-format=1.3] S[table-format=1.3] |
			S[table-format=3.2] S[table-format=3.2] S[table-format=3.2] 
			S[table-format=3.2] S[table-format=3.2]} 
		\toprule
		\multirow{2.5}{*}{\textbf{Resolution}} & 
		\multirow{2.5}{*}{\textbf{GPU}} &
		{\multirow{2.5}{*}{\textbf{\makecell[c]{Time \\ (ms)}}}} &
		{\multirow{2.5}{*}{\textbf{FPS}}} &
		{\multirow{2.5}{*}{\textbf{\makecell[c]{FLOPs \\ (G)}}}} &
		{\multirow{2.5}{*}{\textbf{\makecell[c]{Mem \\ (GB)}}}} &
		{\multirow{2.5}{*}{\textbf{TF}}} &
		{\multirow{2.5}{*}{\textbf{CF}}} &
		{\multirow{2.5}{*}{\textbf{RF}}} &
		\multicolumn{5}{c}{\textbf{EEI Score for Weight Scenario}} \\
		\cmidrule(lr){10-14}
		& & & & & & & & & {\textbf{8:1:1}} & {\textbf{9:0.5:0.5}} & {\textbf{6:2:2}} & {\textbf{9:0:1}} & {\textbf{10:0:0}} \\
		\midrule
		\multirow{3}{*}{1920$\times$1080} 
		& RTX 3090      & 10.48  & 95.45 & 2.48 & 0.719 & 1.032 & 0.023 & 0.304 & 85.83  & 94.51  & 68.45 & 95.92  & 103.20 \\
		& Xavier & 94.54  & 10.58 & 2.48 & 0.686 & 1.075 & 0.023 & 0.291 & 89.10  & 98.27  & 70.75 & 99.61  & 107.45 \\
		& RTX 4060      & 32.20  & 31.06 & 2.48 & 0.695 & 1.094 & 0.023 & 0.295 & 90.69  & 100.04 & 72.00 & 101.40 & 109.39 \\
		\midrule
		\multirow{3}{*}{2560$\times$1440} 
		& RTX 3090      & 18.47  & 54.14 & 4.41 & 1.264 & 1.024 & 0.041 & 0.534 & 87.63  & 94.99  & 72.91 & 97.45  & 102.35 \\
		& Xavier & 170.57 & 5.86  & 4.41 & 1.221 & 1.090 & 0.041 & 0.518 & 92.83  & 100.94 & 76.61 & 103.32 & 109.04 \\
		& RTX 4060      & 58.02  & 17.23 & 4.41 & 1.240 & 1.109 & 0.041 & 0.527 & 94.38  & 102.63 & 77.88 & 105.06 & 110.88 \\
		\midrule
		\multirow{3}{*}{3840$\times$2160} 
		& RTX 3090      & 41.82  & 23.91 & 9.91 & 2.756 & 1.030 & 0.092 & 1.164 & 94.96  & 98.98  & 86.92 & 104.34 & 102.99 \\
		& Xavier & 391.77 & 2.55  & 9.91 & 2.760 & 1.113 & 0.092 & 1.172 & 101.69 & 106.50 & 92.06 & 111.90 & 111.31 \\
		& RTX 4060      & 130.90 & 7.64  & 9.91 & 2.732 & 1.112 & 0.092 & 1.160 & 101.46 & 106.32 & 91.75 & 111.66 & 111.18 \\
		\bottomrule
	\end{tabular}
	\caption{Comparative EEI evaluation of the \texttt{DCE++ light} model across different GPUs (RTX 3090, Jetson Xavier, RTX 4060) and resolutions. Time Factor ($TF$), Complexity Factor ($CF$), and Resource Factor ($RF$) are shown, with a fixed PI Score of 100. EEI scores are presented for various weight scenarios: Default (8:1:1), 9:0.5:0.5, 6:2:2, No CF (9:0:1), and Time Only (10:0:0).}
	\label{tab:eei_combined_gpus}
\end{table}

A key finding from this cross-platform analysis is the remarkable consistency of the EEI score. Despite significant disparities in raw processing capabilities—as evidenced by the large variations in Time (ms) and FPS—the final EEI scores for any given resolution and weighting scenario remain exceptionally stable. For instance, at 1080p with the default 8:1:1 policy, the EEI scores are 85.83 (3090), 89.10 (Xavier), and 90.69 (4060). Specifically, the relative variation in EEI scores across the three platforms is consistently maintained within a 10\% margin for all tested configurations. This demonstrates the robustness of the EEI framework as a hardware-agnostic metric, providing a fair and reliable assessment of a model's intrinsic efficiency, independent of the underlying computational power of the testbed.

\subsection{Overall Conclusion and Implications}
Our comprehensive evaluation demonstrates that \texttt{DCE++ light} is an exceptionally lightweight model with scalable temporal performance. Its primary efficiency bottleneck is memory consumption at resolutions of 4K and higher. The analysis of EEI weighting policies underscores a critical concept: \textbf{model efficiency is not an absolute measure but is context-dependent.} A model's performance profile must be evaluated against the specific priorities—be it latency, memory, or model size—of the target application. The proposed 8:1:1 default policy offers a balanced and practical standard for such evaluations. 

Furthermore, the consistent EEI scores across diverse hardware validate the metric's utility for fair, hardware-agnostic comparisons. For higher-resolution deployments, future work should prioritize memory optimization to address the identified bottleneck.

\section{Architectural Design of the APA Module}

The Adaptive Pre-enhancement Augmentation (APA) module is a computational pre-processing framework designed to transform low-light aerial imagery into a domain more amenable to neural network training. Its architecture is a cascaded, multi-domain pipeline that systematically mitigates common photometric degradations (e.g., low contrast, noise) and chromatic aberrations (e.g., color casts) while preserving perceptual fidelity. The design philosophy is to create a moderately-enhanced yet naturalistic training proxy, thereby simplifying the learning task for the subsequent lightweight network. The process is decomposed into three principal stages, each addressing a specific set of visual artifacts. The mathematical and algorithmic formalization of each stage is detailed below.

\subsection{Stage 1: Edge-Preserving Denoising and Luminance Decomposition}
Low-light aerial images are inherently corrupted by high-gain sensor noise. The initial stage mitigates this by applying an edge-preserving bilateral filter, which smooths homogenous regions while maintaining the integrity of critical structural features. Following denoising, the image is transformed from the BGR color space to the YCrCb color space. This decomposition is crucial as it isolates the luminance component (Y channel) from the chrominance components (Cr, Cb), allowing for targeted luminance adjustments without inducing spurious color shifts.

Let the input image be denoted as $I_{in} \in \mathbb{R}^{H \times W \times 3}$. The initial processing is defined as:
\begin{equation}
	I_{denoised} = \mathcal{F}_{Bilateral}(I_{in}, d, \sigma_{color}, \sigma_{space}),
\end{equation}
where $\mathcal{F}_{Bilateral}$ is the bilateral filtering operator, with $d$ as the pixel neighborhood diameter, and $\sigma_{color}$ and $\sigma_{space}$ as the filter's standard deviations in the color and coordinate spaces, respectively. The subsequent transformation yields the luminance and chrominance channels:
\begin{equation}
	(Y, C_r, C_b) = \mathcal{T}_{BGR \to YCrCb}(I_{denoised}),
\end{equation}
where $\mathcal{T}$ represents the color space transformation.

\subsection{Stage 2: Adaptive Visibility and Contrast Enhancement}
The core of the APA module is its adaptive, non-linear enhancement of the luminance channel. A standard gamma correction uses a fixed exponent, which is suboptimal for images with varying levels of darkness. We introduce an adaptive gamma, $\gamma_{adapt}$, whose value is dynamically computed based on the image's mean luminance. This logarithmic formulation provides a substantial boost to severely underexposed regions while applying a more conservative correction to moderately lit areas, thereby preventing over-enhancement.

The mean luminance, normalized to $[0, 1]$, is first calculated:
\begin{equation}
	\bar{\mu}_Y = \frac{1}{255 \cdot HW} \sum_{i=1}^{H}\sum_{j=1}^{W} Y(i,j).
\end{equation}
The adaptive gamma is then derived as:
\begin{equation}
	\gamma_{adapt} = \text{clip}\left(\gamma_{base} - \lambda \ln(\bar{\mu}_Y + \epsilon), \gamma_{min}, \gamma_{max}\right),
\end{equation}
where $\gamma_{base}$ is the base gamma value, $\lambda$ is a sensitivity parameter (0.5 in our implementation), and $\epsilon$ is a small constant for numerical stability. The result is clipped to a predefined range $[\gamma_{min}, \gamma_{max}]$ to ensure robustness.

The gamma correction is then applied to the luminance channel:
\begin{equation}
	Y_{gamma} = 255 \cdot \left( Y / 255 \right)^{1/\gamma_{adapt}}.
\end{equation}
Following this initial brightness lift, Contrast Limited Adaptive Histogram Equalization (CLAHE) is employed. Its purpose here is not global contrast stretching, but rather the enhancement of local texture and detail within large, uniform dark regions (e.g., unlit fields or rooftops), which are prevalent in aerial perspectives.
\begin{equation}
	Y_{enh} = \mathcal{F}_{CLAHE}(Y_{gamma}).
\end{equation}
The enhanced luminance channel $Y_{enh}$ is then merged back with the original chrominance channels $C_r$ and $C_b$.

\subsection{Stage 3: Perceptual Color and Highlight Correction}
The final stage refines the image's perceptual quality by addressing color casts and potential highlight clipping. Urban night scenes often exhibit a reddish or greenish tint from heterogeneous artificial lighting. To counteract this, we transition to the CIEL*a*b* color space, where the a* channel directly corresponds to the green-red axis. A linear scaling is applied to this channel to neutralize color imbalances by amplifying red tones while suppressing green ones.

First, the BGR image is converted to the CIEL*a*b* space:
\begin{equation}
	(L^*, a^*, b^*) = \mathcal{T}_{BGR \to Lab}(\mathcal{T}_{YCrCb \to BGR}(Y_{enh}, C_r, C_b)).
\end{equation}
The a* channel is then adjusted using a red-boost factor $\alpha_{red}$:
\begin{equation}
	a^*_{adj} = \text{clip}\left((a^* - 128.0) \cdot \alpha_{red} + 128.0, 0, 255\right).
\end{equation}
This operation effectively expands the positive range (reds) and contracts the negative range (greens) relative to the neutral gray point (128).

After converting back to BGR, a final tuning step is performed in the HSV (Hue, Saturation, Value) space. This serves two critical functions: first, a modest saturation boost ($\beta_{sat}$) compensates for color desaturation common in low-light imagery. Second, and more importantly, a highlight suppression factor ($\eta_{supp}$) is applied to the Value (V) channel. This acts as a protective measure to prevent bright light sources (e.g., streetlights) from becoming clipped or over-exposed—a critical failure mode observed in preliminary experiments.
\begin{align}
	I_{tinted} &= \mathcal{T}_{Lab \to BGR}(L^*, a^*_{adj}, b^*), \\
	(H, S, V) &= \mathcal{T}_{BGR \to HSV}(I_{tinted}), \\
	S_{final} &= \text{clip}(S \cdot \beta_{sat}, 0, 255), \\
	V_{final} &= \text{clip}(V \cdot \eta_{supp}, 0, 255).
\end{align}
The final output image $I_{out}$ is obtained by merging the modified HSV channels and converting back to the RGB color space.

\subsection{Algorithmic Summary}

\begin{algorithm}
	\caption{APA: Adaptive Pre-enhancement Augmentation Pipeline}
	\label{alg:apa}
	\begin{algorithmic}[1]
		\Require Input RGB image $I_{in}$, parameters $\Theta = \{d, \sigma_{color}, \sigma_{space}, \gamma_{base}, \lambda, \alpha_{red}, \beta_{sat}, \eta_{supp}\}$
		\Ensure Enhanced RGB image $I_{out}$
		
		\State $I_{bgr} \gets \mathcal{T}_{RGB \to BGR}(I_{in})$
		\State $I_{denoised} \gets \mathcal{F}_{Bilateral}(I_{bgr}, d, \sigma_{color}, \sigma_{space})$ \Comment{Stage 1: Denoising}
		\State $(Y, C_r, C_b) \gets \mathcal{T}_{BGR \to YCrCb}(I_{denoised})$ \Comment{Luminance-Chrominance separation}
		
		\State $\bar{\mu}_Y \gets \text{mean}(Y)/255.0$ \Comment{Stage 2: Adaptive Enhancement}
		\State $\gamma_{adapt} \gets \text{clip}(\gamma_{base} - \lambda \ln(\bar{\mu}_Y + \epsilon), \gamma_{min}, \gamma_{max})$
		\State $Y_{gamma} \gets 255 \cdot (Y/255.0)^{1/\gamma_{adapt}}$
		\State $Y_{enh} \gets \mathcal{F}_{CLAHE}(Y_{gamma})$
		
		\State $I_{enh\_bgr} \gets \mathcal{T}_{YCrCb \to BGR}(Y_{enh}, C_r, C_b)$
		
		\State $(L^*, a^*, b^*) \gets \mathcal{T}_{BGR \to Lab}(I_{enh\_bgr})$ \Comment{Stage 3: Perceptual Correction}
		\State $a^*_{adj} \gets \text{clip}((a^* - 128.0) \cdot \alpha_{red} + 128.0, 0, 255)$
		\State $I_{tinted} \gets \mathcal{T}_{Lab \to BGR}(L^*, a^*_{adj}, b^*)$
		
		\State $(H, S, V) \gets \mathcal{T}_{BGR \to HSV}(I_{tinted})$
		\State $S_{final} \gets \text{clip}(S \cdot \beta_{sat}, 0, 255)$ \Comment{Boost saturation}
		\State $V_{final} \gets \text{clip}(V \cdot \eta_{supp}, 0, 255)$ \Comment{Suppress highlights}
		\State $I_{final\_bgr} \gets \mathcal{T}_{HSV \to BGR}(H, S_{final}, V_{final})$
		
		\State $I_{out} \gets \mathcal{T}_{BGR \to RGB}(I_{final\_bgr})$
		\State \Return $I_{out}$
	\end{algorithmic}
\end{algorithm}

The entire APA pipeline is summarized in Algorithm \ref{alg:apa}. By systematically addressing noise, luminance, local contrast, color casts, and highlights in their respective optimal domains, the APA module transforms challenging raw inputs into a perceptually consistent and more tractable domain for the network to learn from.

\section{Limitations and Failure Cases}
\begin{figure*}[h!]	\centering
	\includegraphics[width=\linewidth]{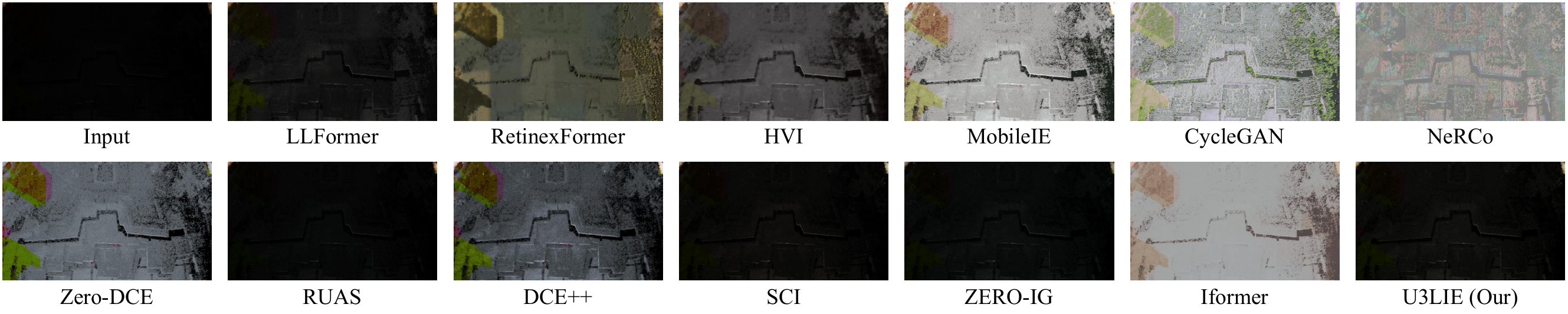} 
	\caption{Illustration of a failure case for U3LIE under extreme low-light conditions.}
	\label{fig:visual_app6}
\end{figure*}

While U3LIE demonstrates robust performance across a wide range of scenes in the U3D dataset, it has limitations. As shown in Figure~\ref{fig:visual_app6}, in scenarios with extreme darkness and virtually no light sources, the framework may struggle to recover meaningful information, sometimes leading to a flat, low-contrast output. This represents a fundamental limitation of any single-image enhancement method when the input signal-to-noise ratio (SNR) falls below a critical threshold, as there is insufficient information to distinguish detail from noise. Future work could explore overcoming this physical boundary by leveraging temporal information from video streams or multi-modal data fusion (e.g., with thermal imagery).

\section{Additional Visualizations}
\begin{figure*}[h!]
	\centering
	\includegraphics[width=\linewidth]{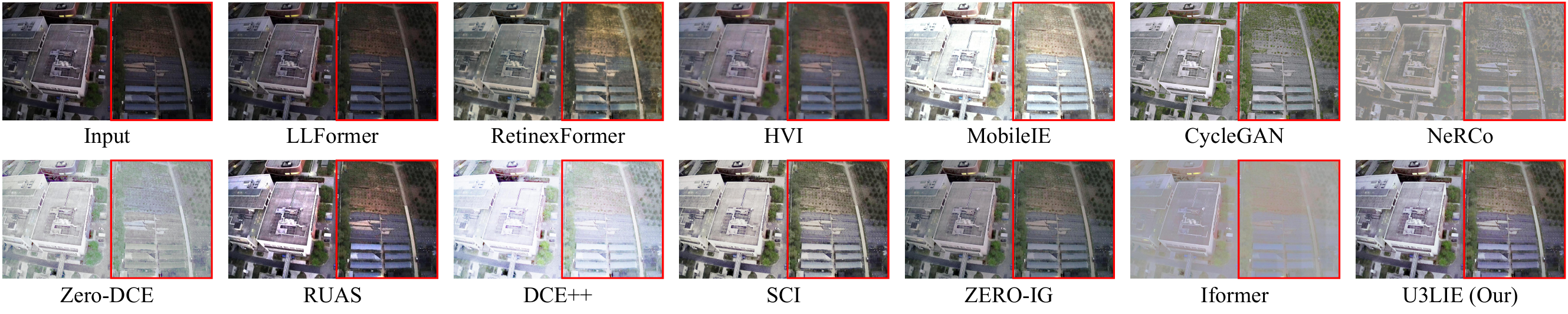} \vspace{3mm}
	
	\includegraphics[width=\linewidth]{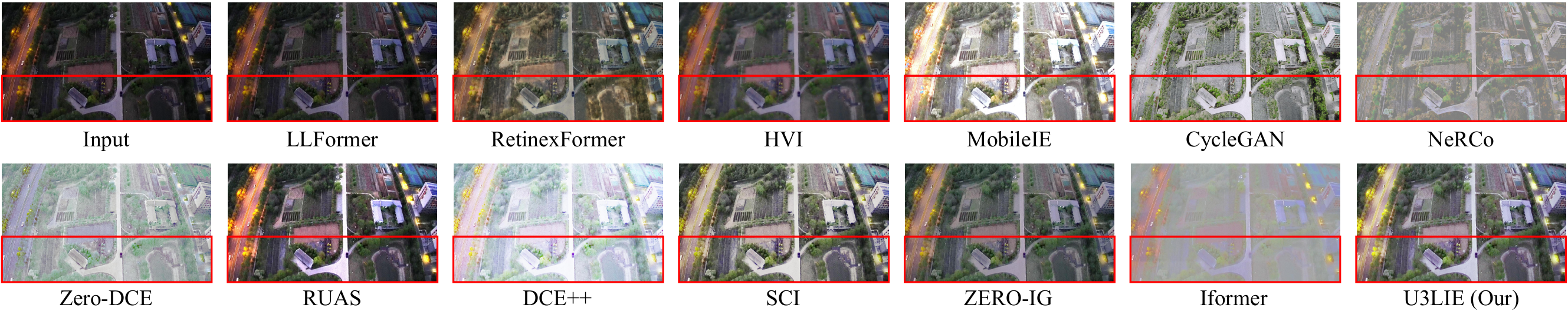} \vspace{3mm}
	
	\includegraphics[width=\linewidth]{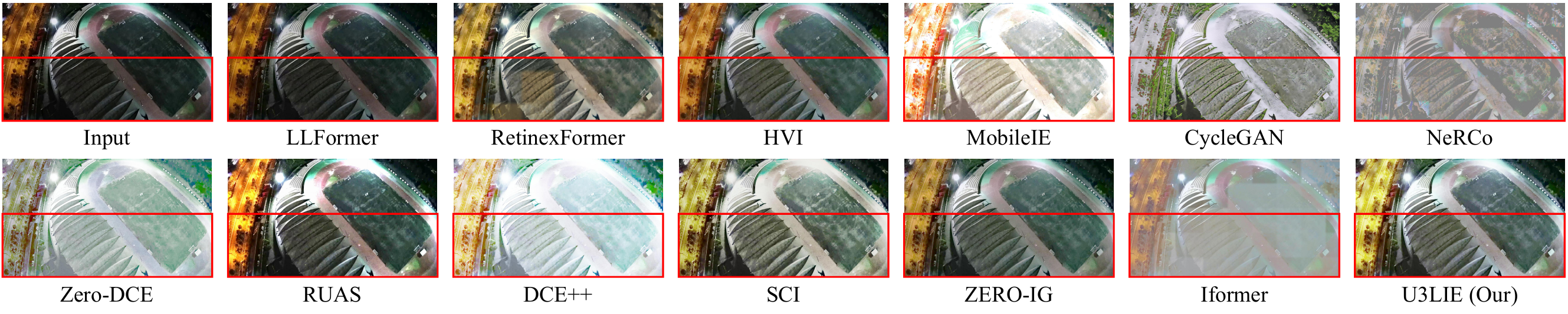} \vspace{3mm}
	
	\includegraphics[width=\linewidth]{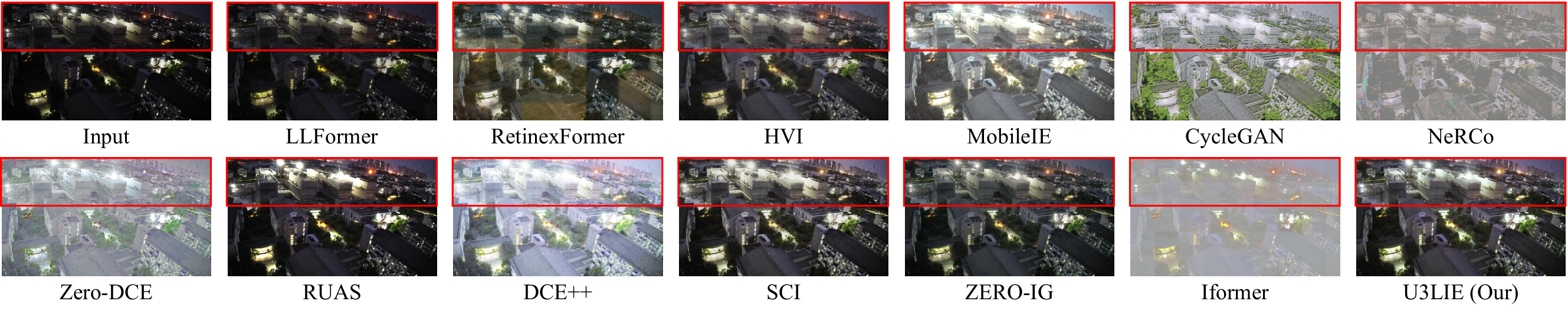} \vspace{3mm}
	
	\includegraphics[width=\linewidth]{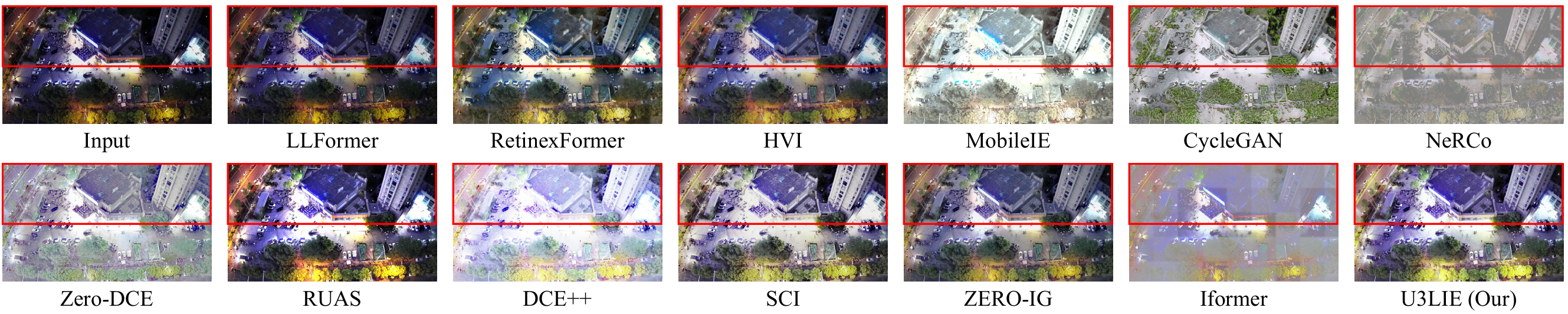} 
	\caption{Visual comparison of our U3LIE with SOTA methods on the U3D dataset.}
	\label{fig:visual_app}
\end{figure*}

This section presents additional visual comparisons to highlight the effectiveness of our approach. Figure~\ref{fig:visual_app} showcases U3LIE's superior capability in handling challenging aerial scenes with non-uniform lighting. It adeptly restores visibility in dark regions while avoiding color artifacts and highlight clipping in bright areas. The insets (red boxes) confirm that our method generates results that are visually plausible and detailed, maintaining the scene's original structure.

\newpage
\bibliography{u3net}

\end{document}